\documentclass{article} 
\usepackage{colm2024_conference}

\usepackage{booktabs}
\usepackage{graphicx}
\usepackage{enumitem}
\usepackage{wrapfig}
\usepackage{algorithm}
\usepackage{algpseudocode}
\usepackage{natbib}
\usepackage{makecell}
\usepackage{booktabs}
\usepackage{array}
\usepackage{amsmath}
\usepackage{amssymb}
\usepackage{amsfonts}
\usepackage{threeparttable}
\usepackage{multirow}
\usepackage{verbatim}
\usepackage{caption}
\usepackage{longtable}
\usepackage{ragged2e}      
\usepackage{array}         
\usepackage{supertabular}
\usepackage{CJKutf8}
\usepackage{array} 
\usepackage[utf8]{inputenc} 
\usepackage[T1]{fontenc}
\usepackage[french,vietnamese,mongolian,greek,english]{babel}
\usepackage{pifont}
\usepackage{xcolor}
\usepackage{enumitem}
\usepackage{tablefootnote}
\usepackage{xspace}
\usepackage{textcomp}
\usepackage{makecell}
\usepackage{lscape} 
\usepackage{siunitx}
\usepackage{listings}
\usepackage{xcolor}
\usepackage{colortbl}
\definecolor{kuaishoublue}{HTML}{6D9EEB}
\hypersetup{
    colorlinks=true,   
}
\definecolor{dt}{gray}{0.7}
\usepackage{pifont}       
\usepackage{bbding}       
\usepackage{fontawesome}
\newcolumntype{L}[1]{>{\raggedright\arraybackslash}m{#1}}

\usepackage{scrextend}

\usepackage{tgpagella}
\usepackage{latexsym}
\usepackage[T1]{fontenc}
\usepackage{microtype}
\definecolor{mydarkblue}{rgb}{0,0.08,0.45}
\definecolor{citecolor}{HTML}{0071BC}
\usepackage{url}            
\usepackage{nicefrac}       
\usepackage{changepage}
\usepackage{xargs}          
\usepackage{wrapfig,lipsum,booktabs}
\usepackage{longtable}
\usepackage{subcaption}
\usepackage{endnotes}

\usepackage{fancyvrb}
\usepackage{fvextra}
\usepackage{pgfplots}
\usetikzlibrary{pgfplots.groupplots}
\pgfplotsset{compat=1.3}
\usepackage{tikz}
\usetikzlibrary{patterns}

\usepackage[most]{tcolorbox}
\usepackage[capitalize,noabbrev]{cleveref}
\crefname{section}{Section}{\S\S}
\Crefname{section}{Section}{\S\S}
\crefname{table}{Table}{Tables}
\crefname{figure}{Figure}{Figures}
\crefname{algorithm}{Algorithm}{}
\crefname{equation}{eq.}{}
\crefname{appendix}{Appendix}{}
\crefformat{section}{Section #2#1#3}
\usepackage{multicol}
\usepackage{tcolorbox}

\usepackage{titlesec}
\titleformat*{\section}{\large\bfseries}

\title{Kwai Keye-VL Technical Report}

\author{
\bf Keye Team, Kuaishou Group}

\begin{document}

\maketitle

\begin{abstract}
While Multimodal Large Language Models (MLLMs) demonstrate remarkable capabilities on static images, they often fall short in comprehending dynamic, information-dense short-form videos, a dominant medium in today's digital landscape. To bridge this gap, we introduce \textbf{Kwai Keye-VL}, an 8-billion-parameter multimodal foundation model engineered for leading-edge performance in short-video understanding while maintaining robust general-purpose vision-language abilities. The development of Keye-VL rests on two core pillars: a massive, high-quality dataset exceeding 600 billion tokens with a strong emphasis on video, and an innovative training recipe. This recipe features a four-stage pre-training process for solid vision-language alignment, followed by a meticulous two-phase post-training process. The first post-training stage enhances foundational capabilities like instruction following, while the second phase focuses on stimulating advanced reasoning. In this second phase, a key innovation is our five-mode ``cold-start'' data mixture, which includes ``thinking'', ``non-thinking'', ``auto-think'', ``think with image'', and high-quality video data. This mixture teaches the model to decide when and how to reason. Subsequent reinforcement learning (RL) and alignment steps further enhance these reasoning capabilities and correct abnormal model behaviors, such as repetitive outputs. To validate our approach, we conduct extensive evaluations, showing that Keye-VL achieves state-of-the-art results on public video benchmarks and remains highly competitive on general image-based tasks (Figure \ref{fig:teaser}). Furthermore, we develop and release the \textbf{KC-MMBench}, a new benchmark tailored for real-world short-video scenarios, where Keye-VL shows a significant advantage. Comprehensive human evaluations also confirm that our model provides a superior user experience compared to other leading models of a similar scale. This paper details the architecture, data construction strategy, and training methodology of Keye-VL, offering valuable insights for building the next generation of MLLMs for the video era.

\end{abstract}

\begin{figure*}[ht]
\centering
\includegraphics[width= 1\linewidth]{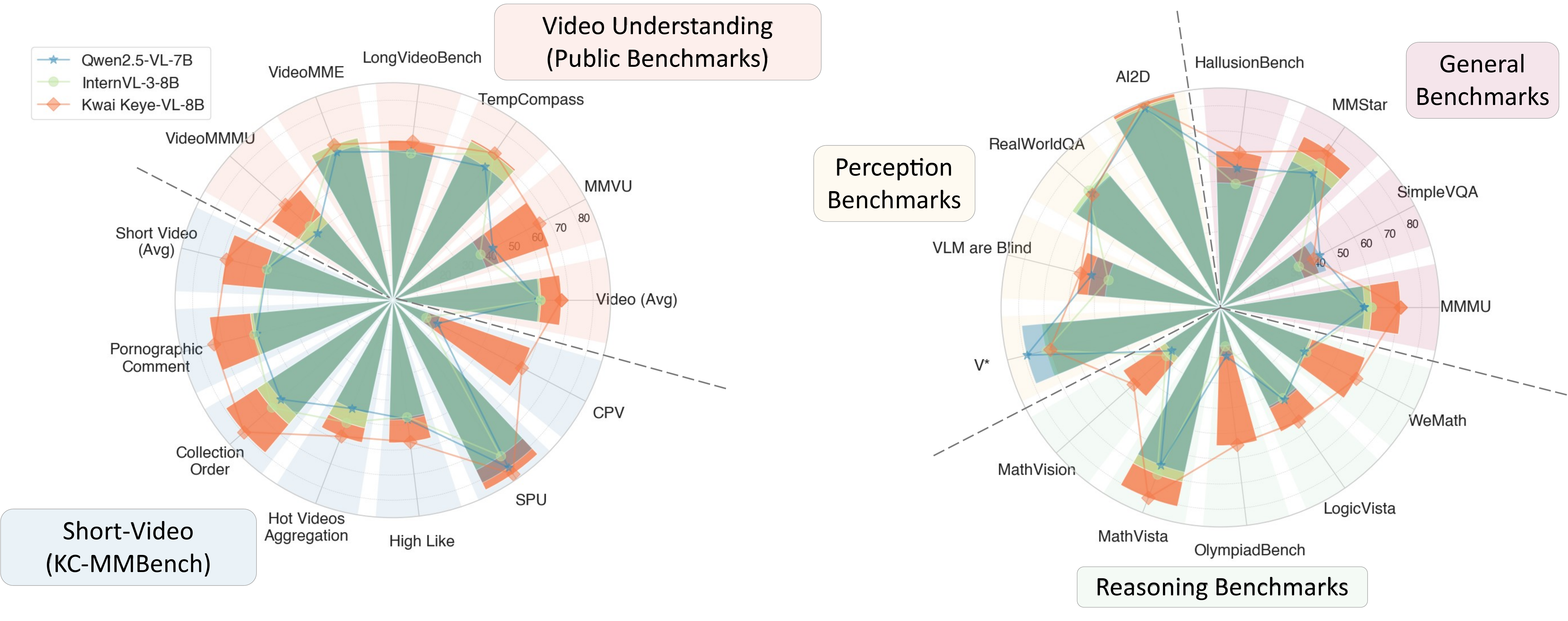}
\caption{\textbf{Benchmark performance of Kwai Keye-VL:} Keye-VL-8B establishes a new state-of-the-art among models of a similar scale, showing a clear lead in video-centric benchmarks (left) while maintaining competitive results on general perception and reasoning tasks (right), validating our training approach.}\label{fig:teaser}
\end{figure*}

\newpage
{
  \setstretch{0.7}
  \tableofcontents
  \noindent\hrulefill
}
\newpage

\section{Introduction}

In recent years, Large Language Models (LLMs) advance rapidly (\cite{llama,phi3,ernie,emu3}), ushering in a new era of artificial intelligence with their powerful capabilities in understanding (\cite{llama4,robo}), generating (\cite{qwen3,seed_1_5}), and reasoning with language (\cite{deepseek_r1, deepseek_v3}). This wave also propels the swift progress of Multimodal Large Language Models (
\cite{o3, mmict,expand,4o,kimi_vl,align_kd,vita_1_5,free,rain,cheap,seed_1_5_vl,seeing,controlm}) (MLLMs), which extend these formidable language abilities to the visual domain, enabling them to perform complex tasks such as visual question answering (\cite{llava,internvl}), detailed image captioning (\cite{feast,eve,li2025denseworld1m}), object grounding (\cite{qwen2_5_vl, mllm_selector}) and visual reasoning (\cite{o3, openthink, open_reasoner}).

Despite significant progress in understanding static images, a major challenge remains in comprehending video content (\cite{long_vita,video_llava}), which is more dynamic and information-dense (\cite{video_rag, gemini_rob}). Short-form videos, in particular, become the primary medium for communication, entertainment, and commerce on platforms like Kuaishou( \cite{onerec,kuaimod}). Understanding short videos is far more complex than recognizing individual objects (\cite{llava}); it requires a model to deeply comprehend the sequence of events, causal relationships, and the overall narrative. Furthermore, the model must integrate information from multiple sources, including video frames and audio (converted to text via Automatic Speech Recognition). Most existing multimodal models, primarily designed for handling combinations of single images and text, lack deep exploration for video tasks and thus often fail to capture the rich, contextual, and sequential information present in videos.
To address this critical gap, we introduce \textbf{Kwai Keye-VL}, a meticulously engineered 8-billion-parameter multimodal foundation model. It achieves leading-edge performance in short-video understanding while also maintaining robust capabilities in general-purpose vision-language tasks (as shown in Figure \ref{fig:teaser}). Our work is driven by a pressing need for a model that not only "sees" the world but also ``thinks'' about its dynamic patterns. This is crucial for enhancing user experience and enabling more intelligent applications in content creation, recommendation, and e-commerce on video-centric platforms.

The development of Kwai Keye-VL rests upon several core techniques. First, we construct a large-scale and diverse dataset exceeding 600 billion tokens, with a special focus on high-quality video data. This data undergoes a rigorous processing pipeline, including filtering, re-captioning with advanced models to generate more precise descriptions, and frame-level annotation to ensure quality. Second, we design an innovative training methodology, which includes a four-stage pre-training process to build a solid foundation for vision-language alignment. Following pre-training, we further enhance Keye-VL's capabilities through a two-phase post-training process:

\begin{itemize}[leftmargin=*, noitemsep]
    \renewcommand\labelitemi{$\diamond$}
    \item \textbf{Stage 1: optimizing foundational capabilities:} We focus on improving the model's basic performance in areas like instruction following. This is achieved through supervised fine-tuning (SFT) and mixed preference optimization (MPO) on high-quality data.
    
    \item \textbf{Stage 2: stimulating and enhancing reasoning abilities:} We begin by creating high-quality "cold-start" data containing five modes: conventional question-answering, long chain-of-thought, auto-reasoning decision, "think with an image" (e.g., generating code to process images), and high-quality video data. We train the model on a mix of these modes, teaching it to select the most appropriate response style. This allows it to think deeply for complex reasoning tasks while responding quickly to simple ones. Subsequently, we employ reinforcement learning to further strengthen its complex reasoning skills. Finally, we use the MPO algorithm for several rounds of iterative alignment to correct issues such as repetitive outputs and flawed logical reasoning.
\end{itemize}

Throughout both pre-training and post-training, we perform rigorous data de-duplication. We compare our training data against general benchmark samples and remove those with high similarity. This process also reveals several public datasets currently implicated in significant data leakage, which we argue should be avoided in model training.

To validate our approach's effectiveness, we first conduct extensive evaluations on multiple public benchmarks. As Figure \ref{fig:teaser} shows, Keye-VL's performance is highly competitive in general image understanding and reasoning, even reaching state-of-the-art level. In video understanding, Keye-VL-8B substantially outperforms previous state-of-the-art models across several benchmarks. To specifically address short-video understanding, we also develop and open-source the Kuaishou Community Multimodal Benchmark (KC-MMBench) and a comprehensive internal evaluation suite to rigorously assess the model's capabilities in realistic, video-centric commercial application scenarios. On this benchmark, Keye-VL-8B also demonstrates a significant advantage, highlighting its value for commercial applications.

For a more granular assessment of Keye-VL's capabilities across various aspects and its real-world user experience, we construct an additional fine-grained internal benchmark. On this benchmark, we conduct a comprehensive and detailed human evaluation of similarly-sized models, including Keye-VL-8B, Qwen2.5-VL-7B, InternVL3-8B, and MiMo-VL-7B. The evaluation results indicate that our model delivers a superior user experience on both video and image-text tasks compared to these baselines. Concurrently, we present an in-depth analysis of Keye-VL's current limitations in fine-grained perception, temporal understanding, and high-level reasoning, outlining future directions for development.

In summary, this paper provides a detailed account of the Kwai Keye-VL's architecture, data processing pipeline, training methodology, and comprehensive evaluation results, offering valuable insights for building the next generation of Multimodal Large Language Models for the video era.

\begin{figure*}[t]
\centering
\includegraphics[width= 1\linewidth]{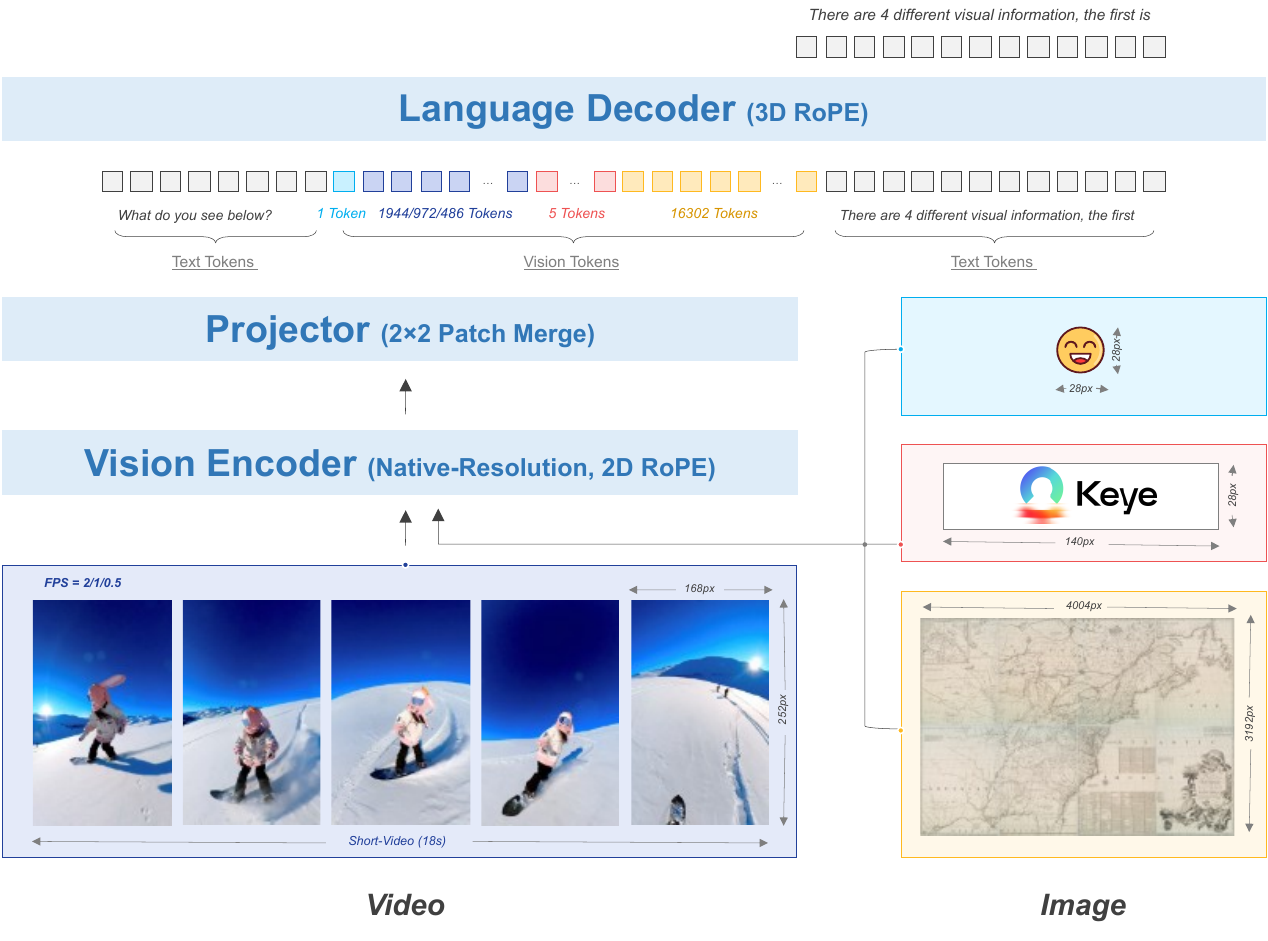}
\caption{\textbf{The Kwai Keye-VL model architecture} is based on the Qwen3-8B language model and incorporates a vision encoder initialized from the open-source SigLIP. It supports native dynamic resolution, preserving the original aspect ratio of images by dividing each into a 14x14 patch sequence. A simple MLP layer then maps and merges the visual tokens. The model uses 3D RoPE for unified processing of text, image, and video information, establishing a one-to-one correspondence between position encoding and absolute time to ensure precise perception of temporal changes in video information.}

\label{fig:arc}
\end{figure*}

\section{Model Architecture}
Figure~\ref{fig:arc}  gives a high-level overview of our Keye-VL, which follows a classic MLLM architecture that includes three key components: a Vision Transformer (ViT), a MLP projector, and a language decoder.
For ViT component, we apply the open-source SigLIP-400M-384-14 \footnote{\url{https://huggingface.co/google/siglip-so400m-patch14-384}} as our vision encoder to extract vision information.
For LLM component, we employ the widely used Qwen3-8B as our language decoder, to provide the universal world semantic knowledge understanding capabilities.
For the projector, we randomly initialize its parameters and fully pre-training it at the Stage 1.
In the following sections, we provide our key upgrades, data pipeline and training recipes.

\subsection{Vision Encoder with Native-Resolution}
In past years, many MLLMs efforts have adopted the well-trained fixed-resolution ViTs as their vision encoders, such as ViT-bigG (\cite{vit_big}
), SigLIP-400M (\cite{siglip}) and others.
However, unlike pre-trained CLIP-based ViTs (\cite{clip}
) that only handle coarse-grained image-caption matching task during training, MLLMs often tackle various finer-grained generation tasks, existing a large gap between them. 
Therefore, we anticipate that our ViT will possess the following capabilities: during processing, images and videos maintain their structural integrity and all details are preserved.

To this end, there are some pioneer MLLMs exploring native-resolution ViT in recent years, such as Qwen2.5-VL, Seed-VL-1.5, Kimi-VL, etc.
In Keye-VL, we also implement a native-resolution ViT, to naturally process images at original resolution, avoiding some complex and redundant image splicing/splitting operations (e.g., MiniCPM2 (\cite{minicpm_v})).
Specifically, our ViT is initialized by the SigLIP-400M-384-14, a fixed-resolution variant with absolute learnable position embeddings to inject the spatial information.
According to it, we first employ interpolation techniques to extend fixed-length learnable position embeddings into resolution-adaptive position embeddings, enabling our basic native-resolution modeling while preserving the pretrained workflow.
Afterwards, to further enhance extrapolation capabilities for positional encoding along visual dimensions, we introduce 2D Rotary Position Embedding (RoPE) to strengthen the visual information modeling. 
In our trial experience, we observe that incorporating 2D RoPE significantly improves the model's performance on high-resolution image.
Finally, building upon the two types of position embeddings, we incorporate the NaViT packing with FlashAttention techniques to continue training our ViT across images with varying resolutions.

During the ViT pre-training procedure, we optimize our native-resolution modifications via SigLIP loss function (the text tower is also from SigLIP-400M-384-14). We use the same distribution data as the downstream MLLM for training, including a total of 500B Tokens from open source data DataComp (\cite{datacomp}), LAION (\cite{laion}), CC12M (\cite{cc12m}), PD12M (\cite{pd12m}), COCO (\cite{coco}) and other in-house data.

\subsection{Visual Encoding}
To guarantee that our language decoder can perceive enough visual signals to understand images and videos in detail, we leave sufficient token buffer for image and videos modeling.

For images of different resolutions, we set the total number of tokens for each image to 16384, which can cover images with more than one million pixels and is sufficient to help the model to see the details of the image in most scenarios.
For video modeling, we devise a dynamic resolution strategy that balances the maximum number of frames and the total number of tokens.
In Keye-VL, we currently set the min/max token number per frame as 128/768, and the max vision token as 24576, this setting can automatically make trade-off between the breadth and depth of visual perception.
Subsequently, based on the extracted frames, we re-calculate the FPS and ensure strict alignment in the  time position in 3D RoPE dimensions during training (position +1 corresponds to +0.5 second in real world).
Meanwhile, we are exploring other more efficient frame modeling techniques to ensure that more frames could feed to our LLM with acceptable computation.

\section{Pre-Training}
In this section, we first describe the construction of the pre-training dataset, followed by an overview of the overall training pipeline and configuration.

\subsection{Data Pipeline}
In our data construction pipeline, we have assembled a diverse, high-quality corpus with exceeding 600 billion tokens to support our models training, sourced from both public datasets and proprietary in-house data.
Generally, our training data encompasses six primary categories: Image Caption, OCR \& VQA, Grounding \& Counting, Interleaved, Video Understanding and Pure Text data.
To ensure these overall data quality, we have designed customized filtering mechanisms tailored to the characteristics of each data category.
For large volumes of medium-quality data, we employ CLIP (\cite{clip}) scores for preliminary filtering. 
For smaller amounts of high-quality data, we utilize open-source MLLMs as discriminators for data selection.
Additionally, we also conduct rigorous image-based deduplication operation, to avoid the potential data leakage between our training corpus and evaluation benchmarks (\cite{corruption}).
Specifically, we identify highly similar images, then remove these near-duplicates from the dataset.
We detail and list a part of deduplication results in Table~\ref{tab:dedup}.
In the following sections, we provide detailed descriptions of each category of training data.

\subsubsection{Image Caption Data}
Image caption task provides the fundamental world knowledge to establish a mapping relationship between visual features and linguistic concepts by pairing image with textual descriptions.
Based on large-scale caption data, our model gains the ability to perceive and comprehend a broad, rich spectrum of world knowledge, such as real-world physical principles and cultural conventions.
Although we can public access many diverse Chinese and English open-source caption data source, such as LAION (\cite{laion}), DataComp (\cite{datacomp}) and Coyo (\cite{coyo}),
the quality of such data is often unreliable, as it typically only undergoes simple crawler-based matching.

To alleviate such data noise, we conduct strict similarity-based filtering pipeline to control the data quality, e.g., scoring the raw rigorous image-caption pair by a CLIP model.
In practice, to ensure data quality, we retain high-similarity image-caption pairs (e.g., CLIP score > 0.9) while leveraging filtered low-quality open-source image data and our in-house image data through a re-captioning pipeline.
During the re-caption, we utilize several MLLMs (Qwen2.5-VL 72B (\cite{qwen2_5_vl}), Tarsier2 (\cite{tarsier2}), GPT-4o (\cite{4o}), Gemini1.5-pro (\cite{gemini}) and others) to generate the synthesis caption for vary resolution images and image category information.
In our experience, we find that recaption data generated by different MLLMs can be very helpful for fine-grained image understanding.

\subsubsection{OCR \& VQA Data}
Optical Character Recognition (OCR) and Visual Question Answering (VQA) are vital tasks to encourage our model to distinguish the details of images.
By integrating OCR capabilities, the model can accurately extract and interpret textual information within images, while VQA task enables our model to comprehend and reason about visual content in a context-aware manner.
In order to build our capabilities in OCR and VQA, we have collected a large number of open-source data, such as Latex-Formula, hand-write text, real-world street views, charts, rich-text documents, multi-image OCR and so on.
Since most of the open-source datasets are in English, to further enhance the model's capability in Chinese OCR \& VQA tasks, we introduce multiple techniques for synthesizing in-house Chinese data:
\begin{itemize}[leftmargin=*, noitemsep]
\renewcommand\labelitemi{$\diamond$}
    \item \textbf{Synthesis:} To enhance the model's OCR capabilities, we aggregate both open-source and in-house image-text datasets, utilizing the text-dense images to build a comprehensive OCR dataset which covering diverse scenario. For VQA task, we first design a set of seed-questions and expand the initial question pool through self-evolution methods. Next, both images and their corresponding captions are fed into MLLMs to generate high-quality and diverse VQA data.
    \item \textbf{Rendering:} Considering the scarcity of high-quality open-source Chinese OCR data, we further leverage font rendering tools to synthesize high-quality OCR samples (includes (1) diverse image backgrounds/layout,  (2) semantic/non-semantic text, (3) multiple fonts styles/sizes and (4) vary image resolutions), which significantly enhances the model’s robustness for Chinese OCR recognition.
\end{itemize}

\subsubsection{Grounding \& Counting Data}
Object grounding is one of the fundamental abilities of MLLMs(~\cite{qwen2_5_vl,seed_1_5}), which enables our model to establish a direct connection between visual information and text semantics.
In Keye-VL training, we primarily utilize three object localization forms: center points, bounding boxes, and polygons.
Their coordinates are strictly typed as integers and normalized to the range [0, 1000) for different resolution images, as shown in the Table~\ref{Grounding}.
In general, we mainly employ the RefCoCo (\cite{refcoco}), VisualGenome (\cite{visual_genome}), TolokaVQA (\cite{toloka_vqa}) as our grounding data source, and the PixMo (\cite{pixmo}) as our counting data source.
To filter the incorrect, missing, or ambiguous annotation grounding data, we utilize the CLIP to select the higher-score points/boxes/polygons as our training data, i.e., extracting the corresponding grounding area from the image to compute its similarity with the target objective text.

\begin{table}[t!]
\footnotesize
\centering
\setlength{\tabcolsep}{0pt}{
\resizebox{0.99\textwidth}{!}{
\begin{tabular}{lc}
\toprule
\multicolumn{2}{c}{\textbf{center points}} \\
\midrule
Example& <|point\_start|>[[x1, y1]]<|point\_end|> \\
Description& The [x1, y1] is the center point of queried objective.\\
\midrule
Example& <|point\_start|>[[x1, y1], [x2, y2]]<|point\_end|> \\
Description& Supporting multiple points for a single queried objective.\\
\midrule
Example& <|object\_ref\_start|>obj<|object\_ref\_end|><|point\_start|>[[x1, y1]]<|point\_end|> \\
Description& The [x1, y1] is the center point of `obj'. \\
\midrule
\multicolumn{2}{c}{\textbf{bounding boxes}} \\
\midrule
Example& <|box\_start|>[[x1, y1, x2, y2]]<|box\_end|> \\
Description& The coordinates [x1, y1]/[x2, y2] denote the top-left and bottom-right point of box of queried objective.\\
\midrule
Example& <|box\_start|>[[x1, y1, x2, y2], [x3, y3, x4, y4]]<|box\_end|> \\
Description& Supporting multiple boxes for a single queried objective.\\
\midrule
Example& <|object\_ref\_start|>obj<|object\_ref\_end|><|box\_start|>[[x1, y1, x2, y2]]<|box\_end|>\\
Description& Detecting the `obj' and its corresponding box.\\
\midrule
Example& <|ocr\_text\_start|>text<|ocr\_text\_end|><|box\_start|>[[x1, y1, x2, y2]]<|box\_end|> \\
Description& Identify the OCR results and its corresponding box.\\
\midrule
\multicolumn{2}{c}{\textbf{polygons}} \\
\midrule
Example& <|object\_ref\_start|>obj<|object\_ref\_end|><|polygon\_start|>[[[x1, y1], [x2, y2], [x3, y3]]]<|polygon\_end|> \\
Description& The coordinates [x1, y1], [x2, y2], ... represent polygon vertices of `obj', which arranged in clockwise order.\\
\midrule
Example& <|ocr\_text\_start|>text<|ocr\_text\_end|><|polygon\_start|>[[[x1, y1], [x2, y2], [x3, y3]]]<|polygon\_end|> \\
Description& Supporting the OCR results.\\
\bottomrule
\end{tabular}
}
}
\caption{Grounding Label Assembling of Keye-VL.}
\label{Grounding}
\end{table}

\subsubsection{Interleaved Text-Image Data}
Instead of the learning task surrounding the single images, we also introduce a large amount of interleaved data to enhance our language decoder's longer multi-modal context modeling ability.
Actually, beyond modeling multi-image correlations, the interleaved data could contribute several critical advantages in pre-training: (1) Preservation of General Knowledge: It contains a wealth of universal knowledge, ensuring that the LLM module's core capabilities are not degraded during training, (2) Enhanced Vision-Language Alignment: By leveraging in-context learning, it helps the model better align visual and semantic signals in language model side, (3) Improved Generalization: The diverse and interleaved nature of the data strengthens the model's ability to reason across modalities and generalize to unseen tasks.
Besides the open-source interleaved data, we also build a large-scale in-house interleaved data generation pipeline.
Specifically, we focus on the two type of raw rich-text documents processing, the academic PDF data and structured knowledge data, especially the Science, Technology, Engineering, and Mathematics (STEM) data.
We collect a substantial amount of academic and knowledge-based PDF/structured data to render the text content into plain text format and insert the corresponding images at their original positions within the text.
In such a process, we conduct rigorous data protection strategies to ensure high-quality outputs. Our pipeline includes: (1) Garbled character recognition: identifying and removing garbled characters, (2) Low-resolution/broken image filtering: ensuring image quality, (3) Text-image similarity validation: ensuring semantic alignment between interleaved image-text.

\subsubsection{Video Data}
As a short-video and live-streaming service provider, the video understanding ability is the most important point of Kwai, such as understanding the video details, generating summaries, and expressing interesting implications.
To reach the goal, our video data are collected from multiple sources, including diverse open-source datasets and a large-scale high-quality in-house video data.
Based on these videos, we conduct the following key pipelines to guarantee our data quality:
\begin{itemize}[noitemsep]
\renewcommand\labelitemi{$\diamond$}
    \item Interleaved video-ASR: For audio signals, we currently use speech-to-text tools (e.g., Qwen2.5-Omni (\cite{qwen2_5_omni})) to recognize them, and then form a interleaved style to connect images and audio to our model.
    \item Video recaption: With (optional) ASR results, we next utilize diverse public MLLMs to generate its caption under different FPS setting, such as 0.5/1/2.
    \item Frame-level OCR annotation: In order to ensure that our model does not miss any details in each frame, we further added a frame-level OCR task.
\end{itemize}
In addition to OCR and video captioning tasks, we have designed a series of reasoning-enhanced tasks to help the model better understand contextual relationships in short videos. These include:
\begin{itemize}[leftmargin=*, noitemsep]
\renewcommand\labelitemi{$\diamond$}
\item Frame-level re-ordering: Given a set of shuffled video frames, our model is required to predict their original chronological order, which enhances its ability to grasp temporal progression and logical flow.
\item Multiple video matching: Provided with a group of related videos and a set of candidate videos, our model is required to identify the most contextually relevant candidate, which refines its understanding of semantic connections across different videos.
\end{itemize}

\begin{figure*}[t]
\centering
\includegraphics[width= 1\linewidth]{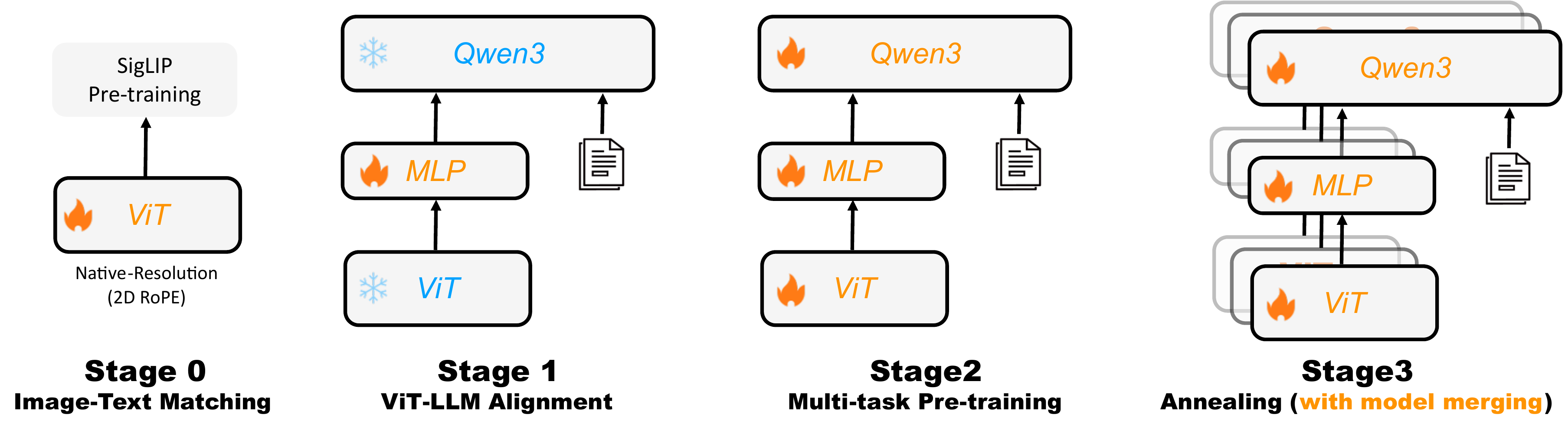}
    \caption{\textbf{The Kwai Keye pre-training pipeline,} featuring a four-stage progressive strategy: Image-Text Matching, ViT-LLM Alignment, Multi-task Pre-training, and Annealing with model merging.}
\label{fig:pretraining_stages}
\end{figure*}

\subsection{Training Recipe}
We employ a four-stage progressive training strategy to build a powerful multi-modal foundation model with strong vision-language alignment capabilities. The training pipeline, illustrated in Figure \ref{fig:pretraining_stages}, is meticulously designed to ensure that each stage has a clear and interconnected objective.

The Vision Transformer (\cite{vit}) (ViT) is initialized with weights from the \textit{siglip-so400m-patch14-384} model and undergoes continuous pre-training using the SigLIP (\cite{siglip}) contrastive loss function. This stage focuses on adapting the vision encoder to our internal data distribution. We incorporate native dynamic resolution processing (akin to NaViT (\cite{navit})), which preserves the original aspect ratio of images to the greatest extent possible. Additionally, 2D Rotary Position Embeddings (\cite{rope}) (RoPE) are integrated to enhance the model's extrapolation capabilities when processing images of varying resolutions.

\paragraph*{Stage 1: cross-modal alignment:}
The language model is initialized from Qwen3-8B (\cite{qwen3}). During this stage, the parameters of both the vision and language models are frozen. Training is focused on optimizing the projection MLP layer. With large-scale datasets, we establish a robust alignment between cross-modal features, laying the groundwork for the subsequent learning phase.

\paragraph*{Stage 2: multi-task pre-training:}
All model parameters are unfrozen for end-to-end optimization using a diverse set of multi-task training data. The data in this stage encompasses a wide range of common vision-language tasks, including Image Captioning, Optical Character Recognition (OCR), Grounding, Visual Question Answering (VQA), and interleaved image-text data. This process significantly enhances the model's fundamental visual understanding capabilities.

\paragraph*{Stage 3: annealing:}
This stage involves an annealing phase where the model is fine-tuned on a curated set of high-quality data. The primary goal is to address the issue of insufficient exposure to high-quality samples during the large-scale, broader training of Stage 2. Through optimized learning strategies and data mixtures, we further refine the model's nuanced understanding and capabilities.

\paragraph{Model merging:}
The performance of pre-trained models on downstream tasks is highly sensitive to the training data mixture, an effect that is particularly pronounced in smaller models (\cite{merge}). Relying on a fixed data ratio selected based on a validation set can amplify the model's intrinsic biases, leading to discrepancies between benchmark performance and real-world application. To alleviate this, in the final phase of pre-training, we explore a homogeneous-heterogeneous merging technique. This involves averaging the weights of models that have been annealed with different data mixtures. This approach preserves the diverse capabilities of the individual models while reducing overall bias and enhancing model robustness.

\section{Post-Training}

\begin{figure*}[t]
\centering
\includegraphics[width= 1\linewidth]{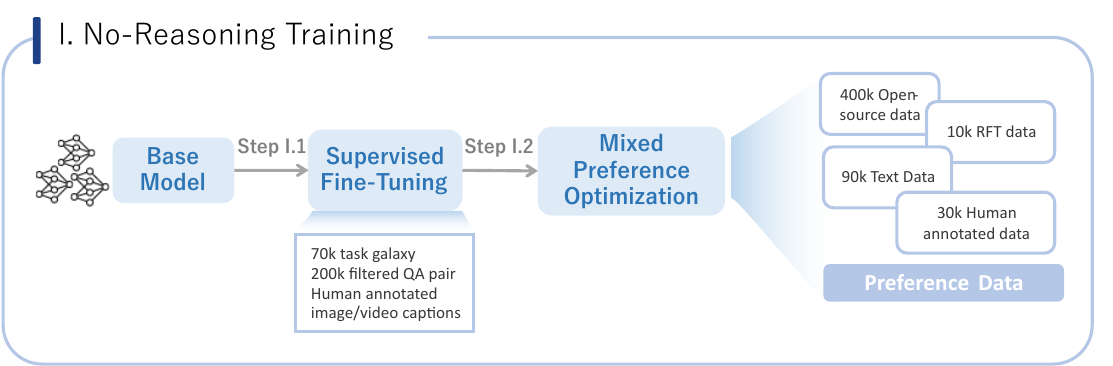}
\caption{\textbf{No-Reasoning Training Pipeline:} The process begins with a Base model, proceeds through Supervised Fine-Tuning (utilizing 70k task galaxy, 200k filtered QA pairs, and human-annotated image/video captions), and culminates in Mixed Preference Optimization with various preference data sources (400k open-source data, 10k RFT data, 90k text data, and 30k human-annotated data).}\label{fig:no-reasoning-training}
\end{figure*}

\begin{figure*}[t]
\centering
\includegraphics[width= 1\linewidth]{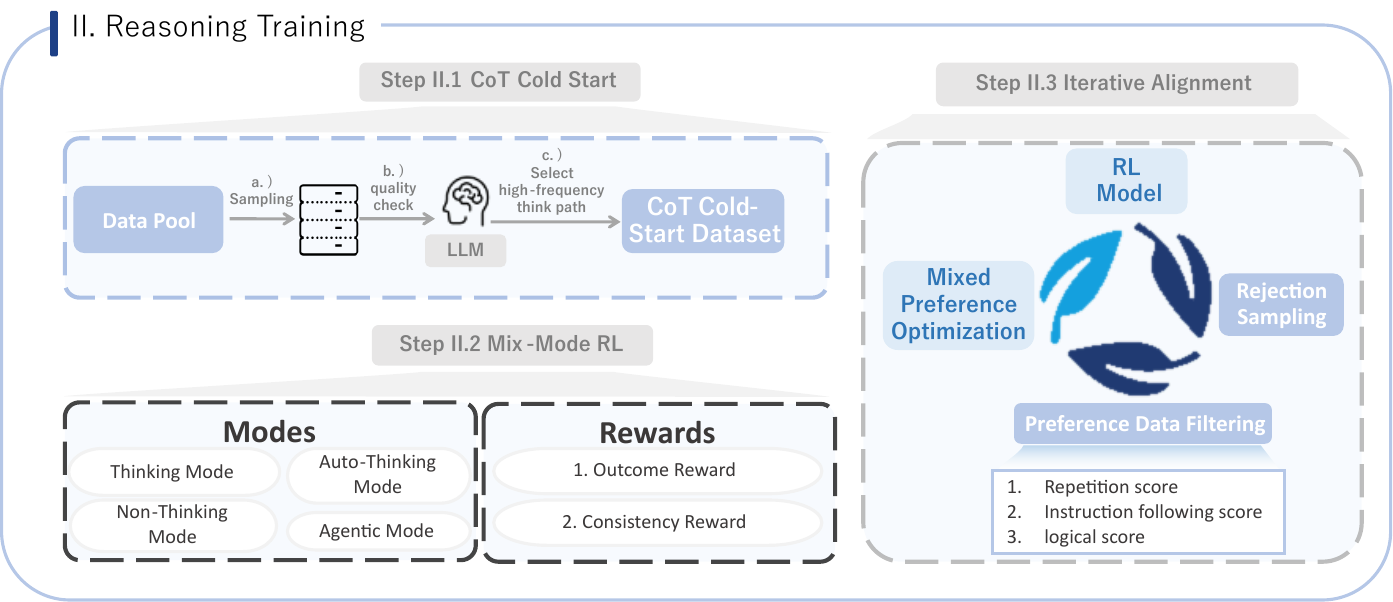}
\caption{\textbf{Reasoning Training Pipeline:} The process consists of three key steps: CoT Cold Start (involving sampling, quality checks, and high-frequency think path selection from a data pool to create a CoT Cold-Start Dataset), Mix-Mode RL (featuring Thinking Mode, Non-Thinking Mode, Auto-Thinking Mode, and Agentic Mode with Outcome and Consistency Rewards), and Iterative Alignment (implementing an RL Model with Mixed Preference Optimization, Rejection Sampling, and Preference Data Filtering based on repetition, instruction following quality, and logical scores).}
\label{fig:reasoning-training}
\end{figure*}

As shown in Figure \ref{fig:no-reasoning-training} and Figure \ref{fig:reasoning-training}, the post-training process for Kwai Keye-VL is a meticulously designed, two-stage methodology engineered to cultivate a comprehensive suite of capabilities. The initial phase, encompassing the first two steps, is dedicated to establishing foundational performance in natural image understanding and text interaction. The subsequent stage, comprising the final three steps, focuses on progressively enhancing the model's sophisticated reasoning abilities.

The training begins with large-scale Supervised Fine-Tuning (SFT) to elevate performance across a wide array of tasks. This is followed by Mixed Preference Optimization (\cite{mpo}) (MPO) to solidify model stability and efficacy in non-reasoning contexts. The third and fourth stages mark a significant leap in cognitive function, introducing Chain-of-Thought (\cite{cot}) (CoT) capabilities through Cold-Start and further refining them via RL. The final stage employs iterative alignment to construct high-quality preference data, which empowers the model to autonomously select the appropriate reasoning mode, thereby ensuring robust and stable performance in practical applications.

To prevent data leakage during post-training, we perform strict data deduplication by removing training samples that are highly similar to common benchmark examples, thereby ensuring fair and unbiased evaluation. Detailed information can be found in \Cref{post_decont}.

\subsection{No-Reasoning Training: Establishing Foundational Performance}

This initial phase establishes the model's core performance and stability in non-reasoning scenarios through two sequential steps (Figure \ref{fig:no-reasoning-training}).

\subsubsection{Step I.1: Supervised Fine-Tuning}

The SFT data candidate pool contains over 5 million multimodal QA samples. We employ the following construction methods to balance comprehensiveness and data quality.

\begin{itemize}[leftmargin=*, noitemsep]
\renewcommand\labelitemi{$\diamond$} 
\item \textbf{To ensure task diversity,} we utilize the proprietary TaskGalaxy (\cite{taskgalaxy}) framework, which categorizes data across a comprehensive system of 70,000 distinct multimodal task types.
\item \textbf{To ensure the data's challenge,} MLLMs are employed to generate multiple reasoning paths for each data point. The complexity of each sample is then measured based on the correctness and length of these responses, allowing for the filtration of overly simple data.
\item \textbf{To ensure data reliability,} human annotators have meticulously crafted captions for the images and videos within the training set.
\end{itemize}
The training strategy involves a dynamic learning rate. In the later phases of training, the model undergoes an annealing process at a lower learning rate. Evaluations show this annealing step contributes approximately a 1\% performance improvement across both open-source and internal benchmarks.

\subsubsection{Step I.2: Mixed Preference Optimization}
Following SFT, the model undergoes MPO to continuously refine its performance.
The dataset composition includes 400,000 open-source samples, 50,000 re-constructed preference samples, 10,000 self-improvement samples, 90,000 text-only samples, and 30,000 human-annotated samples. For open-source data, simple deduplication and filtering of existing multimodal preference data are performed, retaining 400,000 samples. The construction methods for the remaining data are as follows:

\textbf{Re-constructed preference data:} Datasets with ground truth answers and correct responses, such as MM-RLHF (\cite{mm_rlhf}) and MMPR, are collected, and open-source large models (e.g., Qwen2.5-VL 72B) are used to sample high-quality negative examples.

\textbf{Reinforcement fine-tuning (RFT) data:} Preference pairs targeting the SFT model's weaknesses are specifically constructed. The construction process follows two steps: 
(1) Based on benchmark results and human evaluation feedback, tasks where the model performs inadequately are identified, using the SFT target as the chosen results.
(2) Rejection sampling outputs of the SFT model are evaluated using reward models or rule-based rewards to select low-scoring cases as rejected examples.

\textbf{Text-only data:} 90,000 in-house text-only preference pairs are incorporated.

\textbf{Human-annotated data:} Following the MM-RLHF pipeline, different responses to the same prompts are generated using both open-source MLLMs and closed-source APIs. These responses are ranked according to human scoring to create 30,000 human-annotated preference pairs. This high-quality data continuously enhances the performance of this stage.

The training strategy for this stage applies the MPO algorithm, utilizing the constructed paired preference data to optimize its overall performance in non-reasoning contexts.

\subsection{Reasoning Training: Core Breakthrough for Complex Cognition}

This phase represents Kwai Keye-VL's most significant contribution, introducing mix-mode CoT Cold-Start and RL mechanisms to substantially enhance multimodal perception, reasoning, and ``think with image'' capabilities for complex tasks (Figure \ref{fig:reasoning-training}).

\subsubsection{Step II.1: CoT Cold-Start}
\label{cot_mix_cold_start}
This crucial step initializes the model's Chain-of-Thought capabilities. Fine-tuning in this step employs a mixed dataset comprising 330,000 non-reasoning samples, 230,000 reasoning samples, 20,000 automatic reasoning samples, and 100,000 agentic reasoning samples. This strategic balance combines "LongCoT/Thinking" data targeting complex, multi-step reasoning tasks requiring logical rigor and process explainability, with "Instruct/non-thinking" data addressing everyday scenarios demanding quick, clear responses. This combination fosters structured thinking for complex problems while maintaining stylistic diversity and response flexibility for open-ended tasks. The training data consists of:

\textbf{Non-Thinking data/Instruct data:} This dataset mirrors the distribution from SFT and MPO stages without overlapping samples. Our experiments demonstrate that combining Instruct data with Long-CoT data yields performance gains significantly exceeding those achieved by using Long-CoT data alone.

\textbf{Thinking data:} This category includes over 70,000 multimodal samples designed for complex perception and reasoning across mathematics, science, charts, and complex OCR domains. The intuition behind this data construction is to create a ``high-frequency'' or ``commonly traversed'' thinking path for each data point, ensuring accuracy in the long-CoT process. To achieve this, we employ a sophisticated construction process: 1) We use MLLMs to sample multiple CoT paths for each data point. 2) These paths undergo evaluation by another model for correctness in both steps and final results, with only correct samples retained. 3) We identify common, high-repetition thinking steps within these processes and score them based on frequency. 4) Finally, high-frequency thinking processes are identified as complete, high-quality reasoning paths, or high-scoring steps are combined to synthesize new, high-probability paths. The thinking data is supplemented with a long-CoT text dataset generated by sampling responses from another LLM across five domains: code, mathematics, science, instruction following, and reasoning.

\textbf{Auto-Think data:} Sourced from datasets including MMPR, MM-Eureka (\cite{mm_eureka}), and OpenR1-Math (\cite{open_reasoner}), this component trains the model to autonomously determine when to engage its reasoning module. For each prompt, a MLLM first analyzes complexity. Complex prompts trigger responses in a "think + answer" format, while simpler ones receive direct answers. The final training sample comprises the initial analysis, thinking process (for complex cases only), and final answer.

\textbf{Agentic data:} This set of 100,000 samples enables the model's ``think with image'' capability through coding. \textit{Our three-stage construction process creates diverse data that teaches models to manipulate images and perform calculations via code.} First, from 3 million QA pairs, the Qwen 2.5 VL-72B model identifies questions where image operations (cropping, rotation, contrast enhancement) could simplify problems or improve answer quality; the model is then asked to generate corresponding analysis and code. The code will be validated in an external sandbox and the generated image will be checked by another MLLM to ensure code usability. Second, because public datasets rarely contain images requiring rotation or contrast enhancement, to counteract cropping bias, we manually construct low-contrast OCR tasks and artificially rotate images, training the model to write code for contrast enhancement or rotation operations before providing final results, ensuring the model learns more than just cropping and magnification. Third, for mathematical problems, Gemini-2.5-pro generates step-by-step thinking processes, then GPT-4o transforms complex calculations into executable code, broadening the agent's functional scope from image operations to calculation verification. Through these data, the model can perform various image operations (cropping, rotation, magnification, contrast enhancement) via code generation while simultaneously conducting or verifying complex calculations to enhance computational accuracy.

\textbf{Video data:} We have carefully collected a diverse set of video data, encompassing categories such as daily natural videos, movie clips, social media short videos, gaming videos, and educational videos. Among these, there are 24,000 samples labeled as Thinking data and 80,000 samples as Non-Thinking data. The annotation methods and task types can be broadly divided into three categories: (1) various task data directly obtained and filtered from open-source datasets; (2) manually annotated data covering action classification, temporal relationships, and VQA tasks; and (3) tasks constructed using rule-based approaches that focus on temporal modeling—for example, extracting six frames from a video, randomly shuffling their order, and requiring the model to predict the original sequence. These tasks collectively enhance the model's understanding of temporal relationships within videos.

\textbf{Training strategy:} All of the above data are combined during training. Notably, for video samples, we individually sample both 16 and 32 frames per example to participate in training, thereby improving the model's robustness to varying video frame counts. Considering that Long-CoT data primarily targets complex multi-step reasoning tasks with an emphasis on logical inference and process explainability, while Instruct data aligns more closely with everyday usage scenarios that require quick and clear responses, their combination fosters structured thinking for complex tasks while preserving stylistic diversity and flexibility for open-ended tasks. Furthermore, incorporating auto-think and agentic data into mixed training not only boosts model performance in mathematical problem-solving, logical reasoning, and visual perception but also enables flexible output strategy selection according to different user needs. This integration ultimately leads to stronger generalization capabilities and better alignment with human preferences. As a result, the model simultaneously supports the \texttt{/think}, \texttt{/no\_think}, \texttt{/auto\_think}, and \texttt{/agentic\_think} functionalities.

\subsubsection{Step II.2: Mix-Mode RL}
\label{mix_rl}
Building on the CoT cold start, this stage employs RL to further enhance the model's abilities across several key dimensions. The training data is strategically sourced to target specific capabilities:
\begin{itemize}[leftmargin=*, noitemsep]
\renewcommand\labelitemi{$\diamond$}  
\item \textbf{Multimodal perception:} Data involving complex text recognition and object counting is used to maintain the model's perceptual acuity.
\item \textbf{Multimodal reasoning:} Datasets such as MMPR and MM-Eureka are introduced to bolster the model's reasoning capabilities.
\item \textbf{Text-based mathematical reasoning:} The model is challenged with difficult mathematical problems to sharpen its quantitative skills.

\item \textbf{Agentic reasoning:} 47,000 samples from the DeepEyes (\cite{deepeyes}) dataset are incorporated.
\end{itemize}
A Mix-Mode RL strategy using the GRPO (\cite{grpo}) algorithm is applied. The reward signal is provided by large multimodal models, which score both the correctness of the final result and the consistency between the reasoning process and the outcome. 
In addition, we specifically focused on using RL to enhance short video understanding.

\paragraph{RL for short video understanding:} 

At this stage, our goal is to enhance the model’s ability to understand short video content while ensuring broad applicability across diverse video understanding scenarios, which stands out as one of our model’s most distinctive strengths. We aim to enable the model not only to comprehend short video content effectively, but also to assess the video reasonably based on that understanding. By leveraging ground-truth or annotated labels available in various short video content understanding tasks, we apply RL to improve the model’s video reasoning capabilities and align its outputs with the desired value orientations.

\textit{Training procedure.} The RL data for short video understanding is combined proportionally with other data in Step II.2, and the model is trained using GRPO. Compared to the model before RL training, we observe a significant improvement in its performance on short video understanding, with the model’s outputs providing high-quality reasoning paths that align more closely with our expectations across all test tasks. Its assessments of video content better reflect both the judgments of our annotation teams and our intended value orientations.

\subsubsection{Step II.3: Iterative Alignment}
The final step focuses on iterative alignment to address issues like repetitive collapse and flawed reasoning logic. This is achieved using rejection-sampling data sourced from a wide range of domains, including instruction following, OCR, mathematics, charts, counting, text-only content, safety, and cognition. The data construction process involves using the Stage II.2 model to sample multiple responses for each prompt. These responses are then scored and ranked using a hybrid system to create paired MPO data:
\begin{itemize}[leftmargin=*, noitemsep]
\renewcommand\labelitemi{$\diamond$} 
\item \textbf{Rule-based scores:} These metrics assess objective qualities such as repetition and instruction adherence (e.g., format validation, code type verification).
\item \textbf{Model-based scores:} Employing prompt engineering, other MLLMs provide scores for more subjective cognitive aspects of the response.
\end{itemize}
The training strategy consists of multiple iterations on the candidate dataset. In each cycle, ``good cases'' and ``bad cases'' are selected to construct preference pairs, which are then used to update the model via the MPO algorithm. This iterative loop not only refines the model's output but also enhances its ability to assess problem complexity and autonomously select the most appropriate reasoning mode.

\section{Training Infrastructure}
To ensure efficient and stable training of the billion-parameter model, we implement deep optimizations across three key areas: parallelization strategy, load balancing, and fault tolerance.

\paragraph{Optimized hybrid parallelism:} We adopt a hybrid parallelization strategy combining Data Parallelism (DP) and Sequence Parallelism (SP) to scale efficiently across our large compute cluster. Our DP implementation is deeply integrated with the ZeRO (\cite{zero}) optimizer. This not only reduces per-device memory pressure by sharding optimizer states, gradients, and parameters, but more critically, it enables effective computation-communication overlap. During backpropagation, gradient calculation can proceed in parallel with the gradient synchronization communication, effectively hiding communication latency and improving overall training throughput.

\paragraph{Dynamic load balancing:} To address the severe computational load imbalance caused by variable input sizes (images/videos) in multimodal training, we implement a global greedy balancing strategy.  At each global step, this strategy evaluates the FLOPs of each sample in the global batch, sorts all samples in descending order by their FLOPs, and then greedily reassigns them to the parallel group with the current lowest computational load. This mechanism dynamically flattens the load across all nodes, minimizing hardware idle time and significantly boosting overall training speed.

\paragraph{Sample-level auto-resume:} Large-scale training is prone to cause frequent hardware and software failures. To counter this, we build a sample-level auto-resume mechanism. This system performs joint checkpointing of both the training state and the data I/O state. It enables a training job to automatically resume from the exact sample where it was interrupted, requiring no manual intervention. This greatly enhances training stability and resource utilization efficiency.

\paragraph{Post-training framework enhancements:} For post-training, in addition to the strategies above, we update vLLM (\cite{vllm}) to be compatible with Keye's model architecture and video inputs, enabling rapid sampling. Furthermore, we deploy multiple reward models. A random dispatch strategy is employed during the reward calculation process to reduce time overhead during the RL stage.

\section{Evaluation}

\subsection{Zero-shot Image Classification of ViT}
To validate that our continue trained native-resolution ViT is able to capture promising visual representations, we conduct a wide-used zero-shot image classification benchmark analysis.
In our evaluation, we perform a comparative analysis between the base SigLIP model and its two native-resolution position embedding variants, leveraging the CLIP Benchmark\footnote{\url{https://github.com/LAION-AI/CLIP_benchmark}} framework with text prompt template\footnote{\url{https://colab.research.google.com/github/openai/clip/blob/master/notebooks/Prompt_Engineering_for_ImageNet.ipynb\#scrollTo=sRqDoz1Gbsii}}.

The evaluation covers six benchmark datasets: ImageNet-1K, ImageNet-V2, ImageNet-A, ImageNet-R, ImageNet-S and ObjectNet, and its results are shown in Table~\ref{tab:vitzeroshot}.
From it, we have the following observations:
(1) Compared with base SigLIP model, our 1D interpolation position embedding native-resolution model variant has slightly performance degeneration, the reason might be the interpolated 1D position encoding cannot uniquely identify the underlying 2D patch arrangement. For instance, a sequence of 196 patches may correspond to multiple distinct spatial configurations (e.g., 14×14, 7×28, or 28×7), leading to ambiguous spatial localization during feature projection.
(2) With 2D RoPE modification, our ViT could clearly perceive the shape of the image, and showing competitive results with Base SigLIP performance (the best and runner-up results). We think the reason maybe our continued pretraining corpus sharing the same distribution with our MLLMs, rather than the Image-Text matching task.

\begin{table}[t]
\centering
\footnotesize
\begin{threeparttable}
\setlength{\tabcolsep}{1.5mm}{
\resizebox{0.99\textwidth}{!}{
\begin{tabular}{lcccccc} \toprule
    Models & ImageNet-1K & ImageNet-V2 & ImageNet-A & ImageNet-R & ImageNet-S & ObjectNet \\ 
    \midrule
    Base (SigLIP-400M-384-14) & \textbf{83.08} & \textbf{77.34} & \underline{82.22}& \textbf{95.78}& \textbf{74.59} &\underline{76.99} \\
     + 1D interpolation & 82.02	&75.96	&80.92	&94.50	&70.74	&67.58 \\
     + 1D interpolation + 2D RoPE &  \underline{82.65}	&\underline{76.80}	&\textbf{83.26}	&\underline{95.22}	&\underline{72.59}	&\textbf{78.70} \\
    \midrule
\end{tabular}}
}
\end{threeparttable}
\caption{\textbf{Comparison of ViT variants on the ImageNet benchmarks:} The highest scores are marked in \textbf{bold} and the second highest are \underline{underlined}.}
\label{tab:vitzeroshot}
\end{table}

\subsection{Public Benchmarks}

\begin{table}[t]
    \centering
    \small
    \renewcommand{\arraystretch}{1.2}
    \resizebox{0.99\textwidth}{!}{
    \begin{tabular}{l  | S[table-format=2.1] c c | S[table-format=2.1] S[table-format=2.1] S[table-format=2.1] | S[table-format=2.1] S[table-format=2.1]}
    \toprule
    \multirow{2}{*}{\textbf{Benchmark}}  & \textbf{Keye-VL} & \textbf{Keye-VL} & \multirow{2}{*}{} & \textbf{Qwen2.5-VL} & \textbf{InternVL3} & \textbf{MiMo-VL}  & \multicolumn{1}{c}{\multirow{2}{*}{\textbf{GPT-4o}}} & \multicolumn{1}{c}{{\textbf{Claude 3.7}}} \\
    & \textit{8B-Thinking} & \textit{8B-Auto-Think} & & \textit{7B} & \textit{8B} & \textit{7B-RL} &   & \textit{Sonnet} \\
    \midrule
    \rowcolor{kuaishoublue!20}
    \multicolumn{9}{l}{\textbf{General}} \\ %
    MMMU$_{\mathrm{val}}$          & \textbf{71.4}  & \underline{66.8}         &   & 58.6  & 62.7  & {66.7}  &   70.7   & 69.8 \\
    AI2D                                  &  \textbf{86.7} & \underline{}{85.8}        &   & 83.9  & {85.2}  & 83.5  &  82.6 &  81.4 \\
    V*                           & 69.6 &  67.9        &   & \underline{79.1} & {71.2} & \textbf{81.7} & {73.9} & {-} \\
    BLINK$_{\mathrm{val}}$                &  52.0 & 52.5        &   & 56.4  & 55.5  & \textbf{62.4} &  60.0  &  62.3 \\
    VLMs are Blind                        & 57.1 & \underline{61.0}        &   & 37.4 & 36.8 &  \textbf{79.4}     &   49.8 &  72.1 \\
    ZeroBench$_{\mathrm{sub}}$            & \textbf{15.2} & \underline{11.1}         &   & {0.0} & {0.0} & {-} & {-} & {-} \\
    VisuLogic                             & \underline{25.6} & 21.1         &   & {20.0} & \textbf{26.1} & {-} & {-} & {-} \\
    RealWorldQA                           & 67.7 & 66.1         &   & \underline{68.2} & \textbf{70.6} & {-} & {-} & {-} \\
    SimpleVQA                             & \textbf{41.6} & 36.9         &   & \underline{41.4} & {35.1} & {-} & {-} & {-} \\
    MMStar                                & \textbf{75.5} & \underline{72.8}         &   & {64.9} & {68.4} & {70.8} & {-} & {-} \\
    MMVP                                  & \underline{79.0} & \textbf{80.3}         &   & {78.0} & {78.3} & {-} & {-} & {-} \\
    HallusionBench                        & \textbf{67.0} & 57.2         &   & {55.7} & {49.4} & \underline{61.9} & {-} & {-} \\
    All-Angles-Bench                      & 47.3 & \underline{50.3}         &   & {49.4} & \textbf{50.7} & {-} & {-} & {-} \\
    \midrule
    \rowcolor{kuaishoublue!20}
    \multicolumn{9}{l}{\textbf{Doc \& OCR}} \\ %
    ChartQA                               & 86.3 & 72.5        &   & \underline{90.2} & 89.6 & \textbf{91.7}  &  86.7  & 92.2 \\
    CharXiv$_{\mathrm{DQ}}$               &  \underline{77.7} & 75.2        &   & 73.9  & 73.6  &  \textbf{86.8}    &  86.5  &   89.5 \\
    OCRBench                              & 85.1 & 85.3        &   & \textbf{89.7} & \underline{88.0}  &   86.6    & 84.3 &  80.6 \\
    \midrule
    \rowcolor{kuaishoublue!20}
    \multicolumn{9}{l}{\textbf{MATH}} \\ %
    MathVision                           & \underline{46.0} & 42.4       &   & {26.2} & {28.8} & \textbf{60.4} & {31.2} & {-} \\
    MathVista$_{\mathrm{MINI}}$          & \underline{80.7} & 75.2        &   & {66.8} & {70.7} & \textbf{81.5} & {63.8} & {-} \\
    MathVerse$_{\mathrm{vision}}$          & \underline{59.8} & 40.8        &   & {41.2} & {32.4} & \textbf{71.5} & {49.9} & {-} \\
    OlympiadBench                        & \underline{54.8} & 45.2        &   & {19.4} & {25.9} & \textbf{59.4} & {25.9} & {-} \\
    WeMath          & \underline{60.7} & 58.6        &   & {37.7} & {38.5} & \textbf{66.3} & {50.6} & {-} \\
    LogicVista          & \underline{54.8} & 50.6        &   & {44.5} & {43.6} & \textbf{61.4} & {54.4} & {-} \\
    DynaMath          & \underline{37.3} & 35.3       &   & {20.1} & {23.9} & \textbf{45.9} & {54.4} & {-} \\
    \rowcolor{kuaishoublue!20}
    \midrule
    \multicolumn{9}{l}{\textbf{Video}} \\ %
    Video-MME$_{\mathrm{w/o~sub.}}$       & \textbf{67.7} & 59.7         &   & 65.1 & 66.3  &    \underline{67.4}   & 71.9    &     {-} \\
    Video-MMMU                            &  \textbf{57.6}   & \underline{56.9}         &   & 47.4  &   {48.9} &   43.3    & {-}    &    {-} \\
    TempCompass            & \textbf{71.5}  & 58.2         &   & 68.3 &   \underline{70.8} &  68.3   & {-}    &    {-} \\
    LongVideoBench                          &   {62.8} &  \textbf{64.8}         &   & 59.3  &   \underline{63.9}    &    {50.0}   & {-}    &  {-} \\
    MMVU                          &   \textbf{66.1} & \underline{60.3}        &   & {45.5}  &   {39.4}    &    {58.2}   & {-}    &  {-} \\
    \midrule
    \rowcolor{kuaishoublue!20}
    \multicolumn{9}{l}{\textbf{Short-Video}} \\ %
    CPV & \underline{55.1} & \textbf{55.9} &   & {20.1} & 15.0 & 16.7 & {-} & {-} \\
    \makecell[l]{Hot Videos \\ Aggregation} & \underline{54.3} & \textbf{55.0} &   & 46.4 & 52.3 & {49.0} & {-} & {-} \\
    Collection Order & \textbf{84.4} & \underline{82.0} &   & 59.8 & 64.8 & {78.7} & {-} & {-} \\
     \makecell[l]{Pornographic\\ Comment} & \textbf{72.0} & \underline{70.4} &   & 56.1 & 57.1 & {68.3} & {-} & {-} \\
    High Like & \textbf{55.3} & \underline{53.4} &   & 47.9 & 47.0 & {51.1} & {-} & {-} \\
    SPU & \textbf{87.1} & \underline{84.9} &   & 81.3 & 75.6 & {81.9} & {-} & {-} \\
    \bottomrule
    \end{tabular}
    }
    \caption{
        \textbf{Comparison of Keye-VL in \textit{Thinking} and \textit{Auto-Think} mode with other models on diverse visual-language benchmarks:} The best results among open-source models are \textbf{bolded} and the second-best results are \underline{underlined}. 
    }
    \label{tab:general_updated}
\end{table}

In this section, we evaluate Keye-VL across various benchmarks. For \textit{general vision-language tasks}, we select $\text{MMMU}$ (\cite{mmmu}), AI2D (\cite{ai2d}), $\text{V}^\text{*}$ (\cite{vstar}), $\text{BLINK}$ (\cite{blink}), VLMS are Blind (\cite{vlms_blind}), ZeroBench (\cite{zerobench}), VisuLogic (\cite{visu_logic}), RealWorldQA (\cite{real_world_qa}), SimpleVQA (\cite{simple_vqa}), MMStar (\cite{mm_star}), MMVP (\cite{mmvp}), HallusionBench (\cite{hallusionbench}) and All-Angles-Bench (\cite{all_angles_bench}). For \textit{Doc and OCR tasks}, we select ChartQA (\cite{chartqa}), CharXivDQ (\cite{charxivdq}), and OCRBench (\cite{ocrbench}). For \textit{MATH tasks}, we select MathVision (\cite{mathvision}), MathVista\textsubscript{MINI} (\cite{mathvista}), MathVerse\textsubscript{vision} (\cite{mathverse}), OlympiadBench (\cite{olympiadbench}), WeMath (\cite{wemath}), LogicVista (\cite{logicvista}), and DynaMath (\cite{dynamath}). For \textit{public Video tasks}, we select Video-MME(\cite{video_mme}), Video-MMMU (\cite{video_mmmu}), TempCompass (\cite{tempcompass}), LongVideoBench (\cite{longvideobench}), and MMVU (\cite{mmvu}).

 We compare the performance of Keye-VL in \textit{Thinking} and \textit{Auto-Think} mode with state-of-the-art models of a similar scale, including Qwen2.5-VL 7B, InternVL3-8B (\cite{internvl3}), MiMo-VL-7B-RL (\cite{mimovl}), and proprietary models such as GPT-4o and Claude-3.7-Sonnet.

On general vision-language tasks, Keye-VL demonstrates competitive performance across most benchmarks in \textit{Thinking} mode, often achieving SOTA or near SOTA results and outperforming other models overall. On the large-scale general benchmarks $\text{MMMU}_\text{val}$ and AI2D, Keye-VL obtains scores of 71.4\% and 86.7\% respectively, surpassing all other models. On the more challenging $\text{ZeroBench}_\text{sub}$ and MMVP  benchmarks, Keye-VL also achieves the best performance. Furthermore, Keye-VL exhibits a lower hallucination rate, achieving an accuracy of 67\% on HallusionBench. In mathematical reasoning tasks, Keye-VL significantly outperforms Qwen2.5-VL 8B and InternVL3-8B, ranking second only to MiMo-VL 7B-RL. In \textit{Auto-Think} mode, Keye still achieves excellent performance: On $\text{MMMU}_\text{val}$, AI2D, $\text{ZeroBench}_\text{sub}$, and HallusionBench, the performance of Keye-VL in auto-think mode is second only to Keye-VL in thinking mode. On $\text{BLINK}_\text{val}$ and VLMs are Blind, the auto-think mode even surpasses the thinking mode. On MMVP, the auto-think mode achieves state-of-the-art results. On the remaining benchmarks, the auto-think mode results in only a slight performance decline. These results demonstrate Keye-VL's ability to spontaneously select the correct thinking mode and the potential of the auto-think mode to alleviate the phenomenon of over-thinking.

In video-centric scenarios, Keye-VL demonstrates superior capabilities compared to other open-source models. Our evaluations on both public and internal benchmarks indicate that an accurate understanding of video content is one of Keye-VL's core strengths. On public video benchmarks, Keye-VL  significantly outperforms other models in both modes, particularly on Video-MMMU, with an absolute improvement of 8.7\% in thinking mode. On LongVideoBench, auto-think mode surpasses thinking mode by 2\%, this indicates that Keye-VL has sufficiently extracted video information on the input side, allowing it to obtain the correct answer without excessive reasoning.

To better evaluate the short-form video understanding capabilities of Keye-VL, we construct and open-source the Kuaishou Community Multimodal Benchmark (KC-MMBench in Table~\ref{kc_mmmbench})\footnote{The detailed construction process of KC-MMbench is shown in Appendix~\ref{app_kcmmbench}.}. This benchmark assesses model performance across several key dimensions of short-form video comprehension and has undergone rigorous manual inspection and data anonymization. On KC-MMBench, Keye-VL achieves an average accuracy of 68.03\%, substantially surpassing the second-best model, MiMo-VL 7B-RL, which attains an accuracy of 57.62\%. This highlights the effectiveness and application potential of Keye-VL in the domain of short-form video understanding.

\begin{small}
\begin{table*}[t]
    \centering
    \tiny
\resizebox{0.99\textwidth}{!}{
    \begin{tabular}{L{1.5cm} L{2cm} L{2cm} L{5.5cm} L{2.0cm}}
    \toprule
            \multicolumn{1}{c}{~} & \multicolumn{1}{c}{Task Definition} & \multicolumn{1}{c}{Image/Video} & \multicolumn{1}{c}{Input} & \multicolumn{1}{c}{Output}\\ 
    \midrule
        \textbf{\textit{Pornographic Comment}}& The task of justifying whether short video comments contain pornographic content.
        & \begin{minipage}[c]{0.4\textwidth}
    \includegraphics[width=0.3\linewidth]{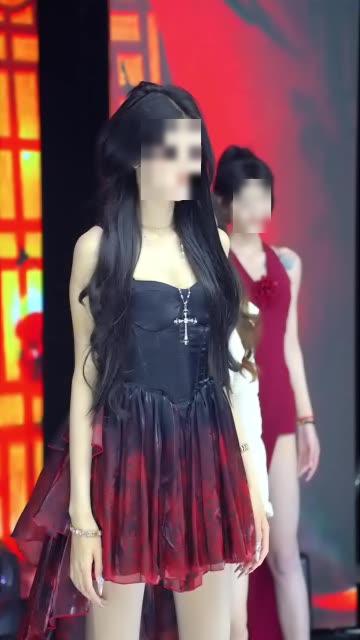} 
\end{minipage} & Please answer the question based on the video cover and extracted video information. The video information is as follows: Video Title: The wind from south of the Yangtze River blows on the birth of all things. Video ASR: You say those sleepless nights and white dreams are hazy and hypoxic... 
User Comment: Every time I watch her dance, I get a special feeling, so good-looking, a queenly vibe [Let's party together].
Please determine whether the comment is pornographic based on the information above. The final output should be "Yes" or "No". & No \\ 
    \midrule
        \textbf{\textit{Collection Order}} & The task of determining the logical order between multiple videos with the same topic. &\begin{minipage}[c]{0.4\textwidth}
    \includegraphics[width=0.3\linewidth]{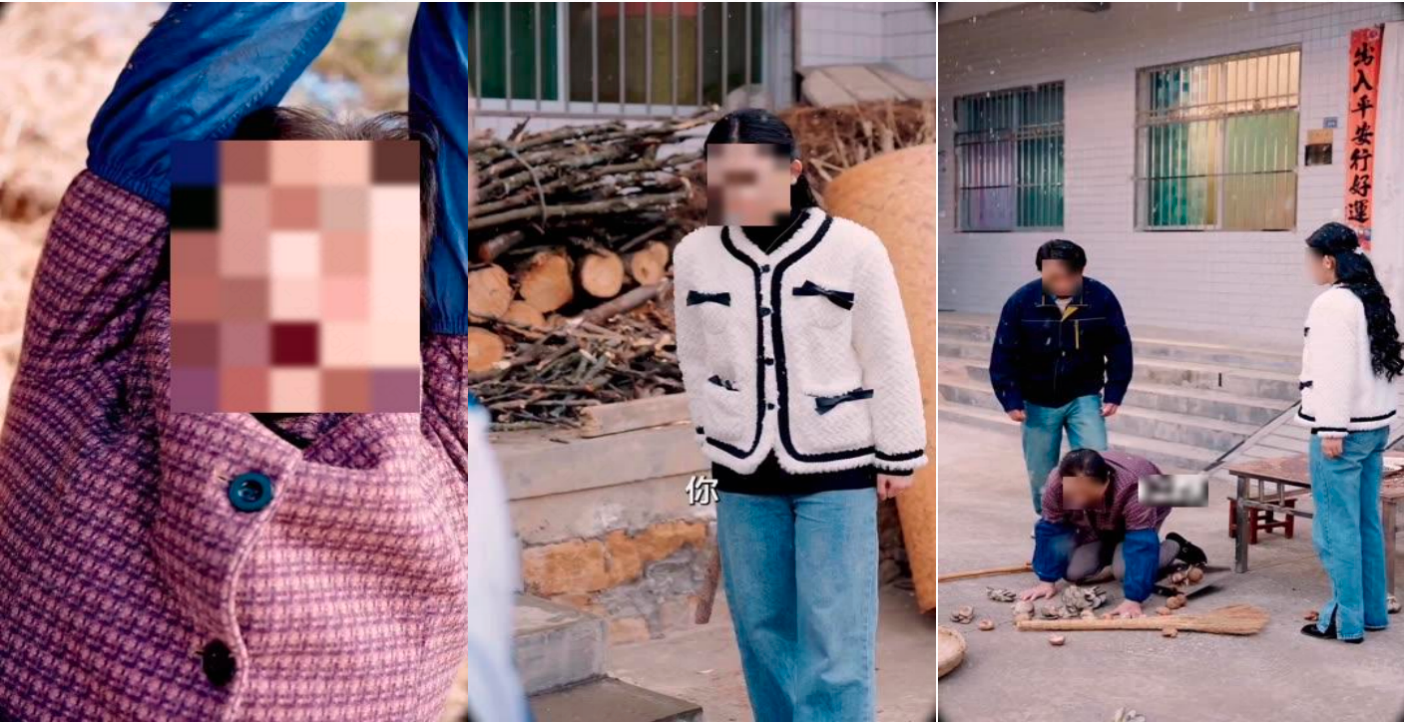} 
\end{minipage} & The following is a list of information for 3 videos. It is known that the 3 videos are from the same video collection. Please sort them based on the video content. Please output your answer in the format of a list, with the video numbers in sequential order, such as: [Video 1, Video 3, Video 2]. &  [Video 3, Video 2, Video 1]\\ 
    \midrule
         \textbf{\textit{Hot Videos Aggregation}} & The task of determining whether multiple videos belong to the same topic. & \begin{minipage}[c]{0.4\textwidth}
    \includegraphics[width=0.3\linewidth]{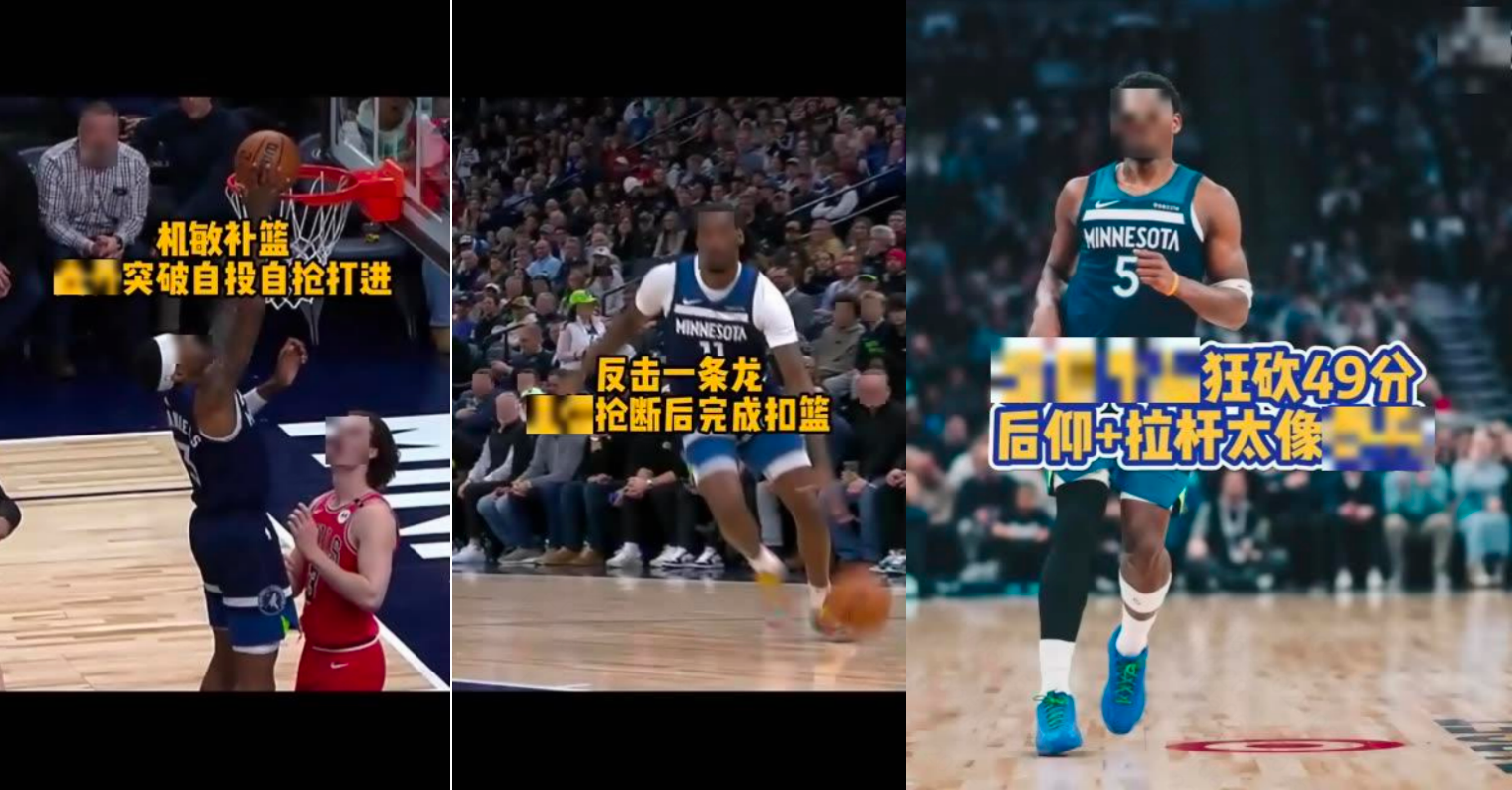}
\end{minipage} & 
The following is a list of information for 3 videos. Please determine if the videos in the given list belong to the same topic as the first video. Each image corresponds to a screenshot of a different video. Please output your answer in the format of a list. Each item in the list represents the judgment of whether the corresponding video belongs to the same topic as the first video. The first video also needs to be included in the list. Example: Assuming there are 3 videos in the list, the output should be "Yes Yes No", which means the second video belongs to the same topic as the first video, and the third video does not belong to the same topic as the first video. & Yes Yes No \\ 

    \midrule
        \textbf{\textit{High Like}} & A binary classification task to determine the rate of likes of a short video. & \begin{minipage}[c]{0.4\textwidth}
    \includegraphics[width=0.3\linewidth]{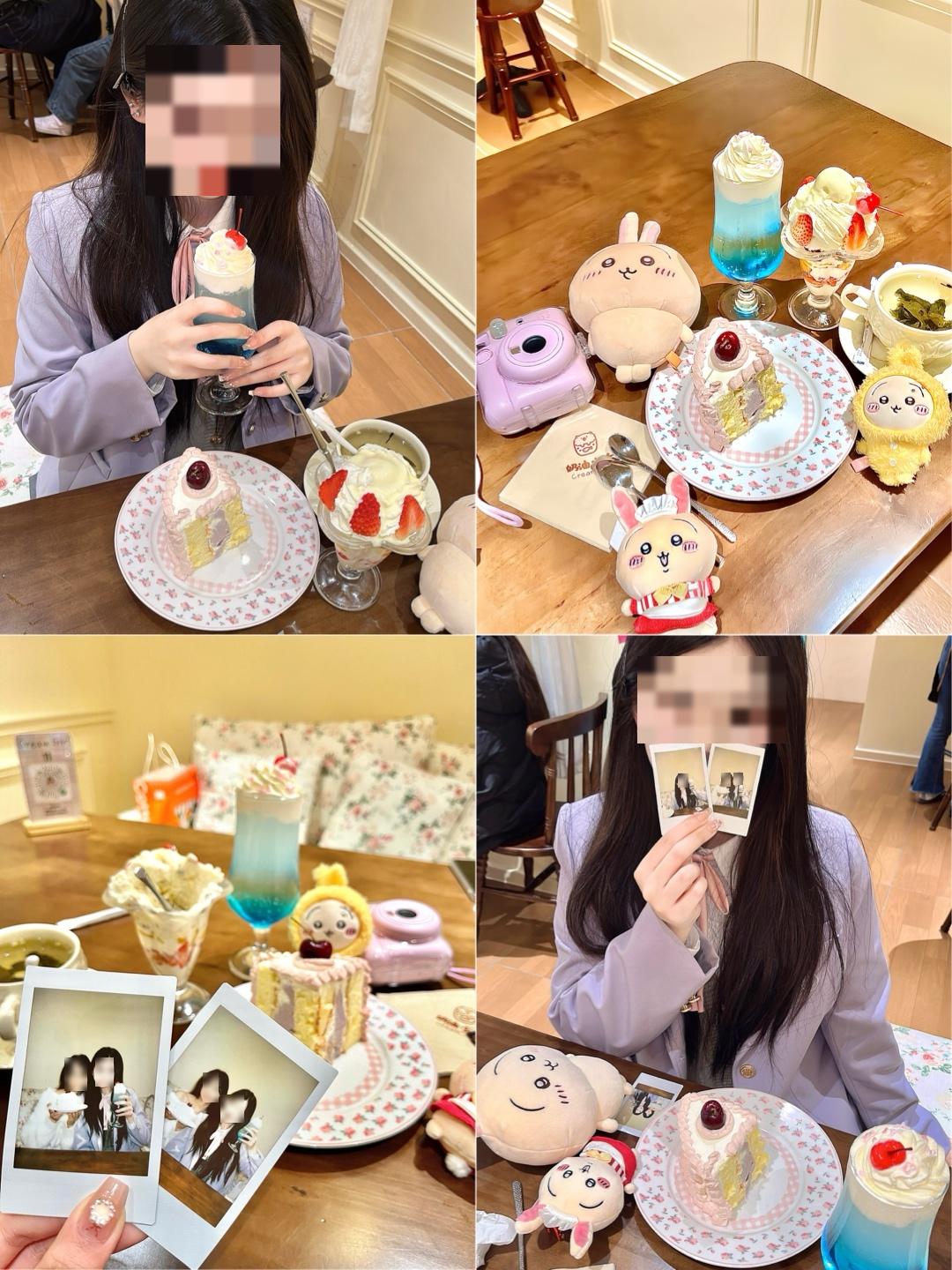} 
\end{minipage} &
Please answer the question based on the video cover and extracted video information. This is a video on a short video platform. Please judge whether the video can get a high like rate on the platform based on the above video content information. The final output needs to be "Yes" or "No". If it can get a high like rate, output "Yes", otherwise output "No". & Yes \\ 

    \midrule
        \textbf{\textit{SPU} (Standard Product Unit)} & The task of determining whether two items are the same product in e-commerce. & \begin{minipage}[c]{0.4\textwidth}
    \includegraphics[width=0.3\linewidth]{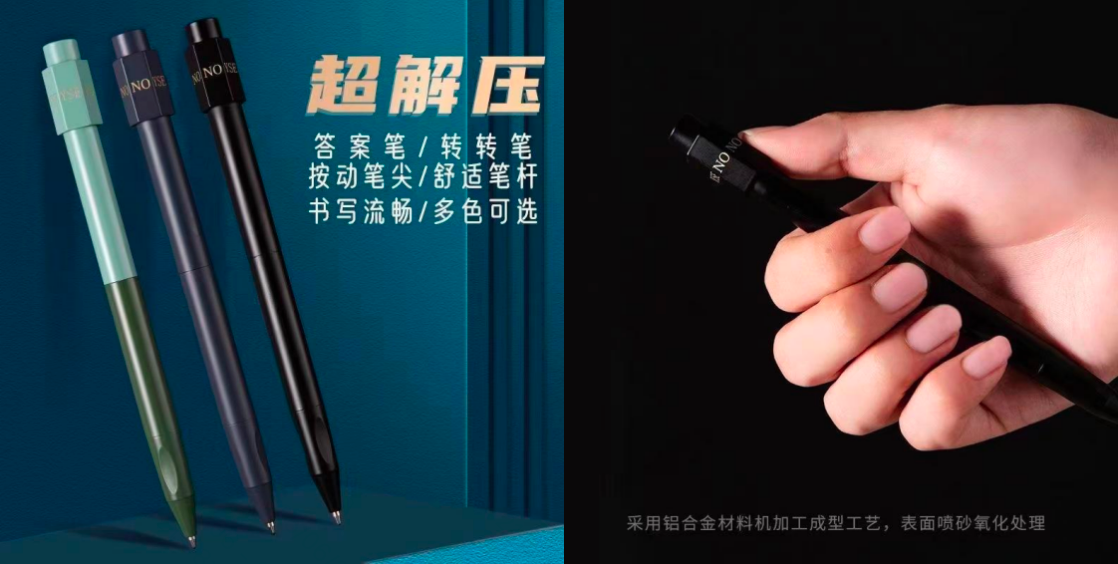} 
\end{minipage} & 
Please determine whether the two given images belong to the same product. & Yes \\ 
    \midrule
        \textbf{\textit{CPV} (Category Property Value)} &  The task of predicting product attributes in e-commerce. & \begin{minipage}[c]{0.4\textwidth}
    \includegraphics[width=0.3\linewidth]{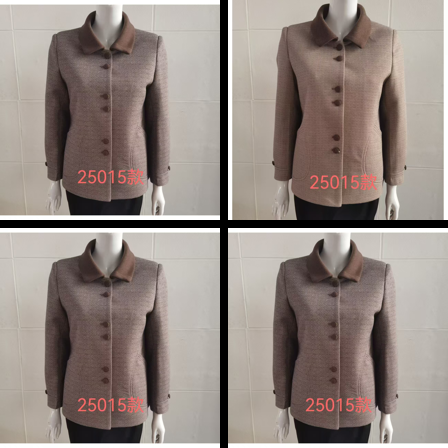}
    \end{minipage} & 
You are an e-commerce AI assistant equipped with a range of capabilities to support various e-commerce operations.
There is now an attribute prediction task. You need to summarize the attributes of the product from the specified dimensions based on the main product image and information (product title and product description) provided to you.
The following is the relevant information of the product whose attributes need to be predicted: Product Title: [Return shipping compensation] 25015 women's new style fashionable jacket [small grid] BY Product Description: None.
Please summarize the attribute information of this product from the dimensions of {'Women's clothing for middle-aged and elderly': ['Small camisole/small sling', 'Jacket/coat'...], 'Sleeve type': ['Regular', 'Petal sleeve'...], 'Whether there is a fur collar': ..., 'Placket':...}, answer in Chinese, and the final attribute summary should be output in json string format. & \{'Women's clothing for middle-aged and elderly': 'Jacket/coat', 'Sleeve type': 'Regular', 'Whether there is a fur collar': 'No', 'Placket': 'Single-breasted', 'Sleeve length': 'Long sleeve', 'Clothing length': 'Regular style'\} \\ 
    \bottomrule
    \end{tabular}
}
\caption{\textbf{Kuaishou Community Multimodal Benchmark}: KC-MMBench is a self-constructed short-video benchmark for MLLMs, which contains 6 task categories and 1840 instances.}\label{kc_mmmbench}

\end{table*}
\end{small}

\subsection{Internal Benchmarks}
Despite extensive evaluations on a wide array of public benchmarks, these benchmarks exhibit numerous limitations that necessitate a focused effort on developing a proprietary, internal evaluation suite. The primary issues are as follows:


\begin{table}[t]
\centering
\footnotesize
\begin{threeparttable}
\setlength{\tabcolsep}{1.5mm}{
\begin{tabular}{llc|ccc} \toprule
    \multirow[t]{2}{*}{\textbf{Subset}} & \multirow[t]{2}{*}{\textbf{Capability (Level 2)}} & \textbf{Keye-VL-8B} & \textbf{Qwen2.5-VL-7B} & \textbf{InternVL3-8B} & \textbf{MiMo-VL-7B} \\
    & & {\scriptsize thinking} & {\scriptsize non-thinking} & {\scriptsize non-thinking} & {\scriptsize thinking} \\
    \midrule
    \multirow{6}{*}{\makecell[l]{Video\\Subset}} & Overall (Average) & \textbf{3.33} & \underline{3.31} & 2.72 & 2.75 \\
    & Correctness & \underline{3.34} & \textbf{3.41} & 2.87 & 3.07 \\
    & Comprehensiveness & \textbf{4.36} & \underline{3.93} & 3.32 & 3.43 \\
    & Relevance & \underline{4.83} & \textbf{4.85} & 4.59 & 4.43 \\
    & Fluency & \textbf{4.89} & \textbf{4.89} & 4.83 & 4.79 \\
    & Creativity & \textbf{3.75} & \underline{3.50} & 2.75 & 2.92 \\
    \midrule
    \multirow{6}{*}{\makecell[l]{Image\\Subset}} & Overall (Average) & \textbf{3.81} & 3.69 & 3.67 & \underline{3.71} \\
    & Correctness & \textbf{4.05} & 3.82 & \underline{3.91} & 3.87 \\
    & Comprehensiveness & \textbf{4.49} & 4.43 & 4.40 & \textbf{4.49} \\
    & Relevance & \underline{4.91} & 4.87 & \textbf{4.93} & 4.74 \\
    & Fluency & \underline{4.89} & \textbf{4.94} & 4.87 & 4.72 \\
    & Creativity & \underline{3.69} & \textbf{4.06} & \underline{3.69} & 3.81 \\
    \bottomrule
\end{tabular}}
\end{threeparttable}
\caption{\textbf{Comparison of Keye-VL with other models on the internal benchmark:} The evaluation is based on human annotations across five dimensions: correctness, comprehensiveness, relevance, fluency, and creativity. The highest scores are marked in \textbf{bold} and the second highest are \underline{underlined}.}
\label{tab:multimodal_eval}
\end{table}

\begin{table}[t]
\centering
\footnotesize
\begin{threeparttable}
\setlength{\tabcolsep}{1.5mm}{
\resizebox{0.99\textwidth}{!}{
\begin{tabular}{ll|cccc} \toprule
    \multirow[t]{2}{*}{\textbf{Subset}} & \multirow[t]{2}{*}{\textbf{Capability (Level 2)}} & \textbf{Keye-VL-8B} & \textbf{Qwen2.5-VL-7B} & \textbf{InternVL3-8B} & \textbf{MiMo-VL-7B} \\
    & & {\scriptsize thinking} & {\scriptsize non-thinking} & {\scriptsize non-thinking} & {\scriptsize thinking} \\
    \midrule
    \multirow{9}{*}{\makecell[l]{Video\\Subset}} & Overall (Average) & \textbf{3.33} & \underline{3.31} & 2.72 & 2.75 \\
    & Visual Element Recognition & \textbf{3.89} & \underline{3.54} & 3.40 & 3.46 \\
    & Temporal Information Understanding & \underline{2.92} & \textbf{2.96} & 2.46 & 2.25 \\
    & Description Ability & \underline{3.08} & \textbf{3.58} & 2.92 & 2.58 \\
    & Creative Ability & \underline{3.17} & \textbf{3.50} & 1.83 & 2.67 \\
    & Knowledge-based QA & \textbf{2.78} & 2.00 & \underline{2.11} & 2.17 \\
    & Reasoning Ability & \underline{3.31} & \textbf{3.50} & 2.85 & 2.88 \\
    & Domain-specific Expertise & \underline{3.36} & \textbf{4.40} & 3.30 & 3.09 \\
    & Robustness & \textbf{3.50} & \underline{3.42} & 2.08 & 2.17 \\
    \midrule
    \multirow{6}{*}{\makecell[l]{Image\\Subset}} & Overall (Average) & \textbf{3.81} & 3.69 & 3.67 & \underline{3.71} \\
    & Visual Recognition & \underline{3.97} & 3.91 & 3.96 & \textbf{3.98} \\
    & Visual Understanding & \underline{3.70} & 3.27 & 3.37 & \textbf{3.73} \\
    & Basic Description & \textbf{4.00} & \underline{3.91} & 3.68 & 3.63 \\
    & Visual Storytelling & \underline{3.63} & \textbf{3.94} & 3.75 & 3.31 \\
    & Multi-image Analysis & \textbf{3.15} & 2.77 & 2.69 & \underline{2.92} \\
    \bottomrule
\end{tabular}}
}
\end{threeparttable}
\caption{\textbf{Detailed capability comparison of Keye-VL with other models on the internal benchmark:} The evaluation examines specific capabilities ranging from basic recognition to complex reasoning and multi-image analysis. The highest scores are marked in \textbf{bold} and the second highest are \underline{underlined}.}
\label{tab:detailed_capability_eval}
\end{table}

\begin{itemize}[leftmargin=*, noitemsep]
\renewcommand\labelitemi{$\diamond$}
    \item \textbf{Benchmark contamination:} An unavoidable limitation of public datasets is the potential for their data to have been inadvertently or deliberately exposed during the model's training process. This phenomenon, known as \textit{benchmark contamination}, can lead to exaggerated performance metrics, reducing the sensitivity of public benchmarks to subtle model improvements and thus failing to reflect the model's capabilities. To obtain a reliable assessment of a model's genuine performance, it is imperative to construct an internal evaluation benchmark that is insulated from training data contamination.

    \item \textbf{Limited language coverage:} The majority of public benchmarks are predominantly focused on English-language scenarios. This significantly constrains the exploration of a model's abilities within the context of native Chinese applications. English benchmarks are incapable of adequately covering or measuring the unique expressions, profound cultural contexts, and diverse local needs inherent to the Chinese language environment. Therefore, building a benchmark that can effectively evaluate a model's Chinese language capabilities is of paramount importance.

    \item \textbf{Insufficient task and domain coverage:} Existing general-purpose evaluation benchmarks primarily concentrate on fundamental perception and simple reasoning abilities. They fail to comprehensively cover a multi-dimensional spectrum of capabilities, including \textit{fine-grained perception}, \textit{cross-modal grounding}, \textit{language modeling}, \textit{complex reasoning}, and \textit{safety \& robustness}. Furthermore, current benchmarks lack a focus on real-world application scenarios, particularly core business needs such as \textit{multimodal understanding in short video communities}. This deficiency results in an evaluation framework that cannot effectively reflect a model's performance in practical business tasks, severely hindering the assessment of its practical value and potential for deployment.

    \item \textbf{Monotonous task difficulty and evaluation format:} The task types and question formats (e.g., true/false, multiple-choice, fixed-answer questions) in existing benchmarks are relatively simple and uniform. They are insufficient for comprehensively measuring a model's capacity to handle complex, \textit{open-ended question answering}, which more closely mirrors authentic user interactions. Constructing a more challenging evaluation benchmark that supports open-ended generative responses is better suited to simulate real-world user and business interactions, thereby enabling a more accurate assessment of the model's generalization capabilities in practical scenarios.
\end{itemize}

\subsubsection{Design Strategies and Core Principles}

In response to the limitations inherent in existing public evaluation benchmarks and motivated by the aforementioned needs, we construct an internal evaluation benchmark set, and the full breakdown for the benchmark is detailed in Table~\ref{tab:general_benchmark}. This structure is intended to fully reflect the model's performance and application potential across different sub-scenarios. While ensuring robust coverage of textual abilities, we place a special emphasis on multimodal capabilities, particularly for images and videos. In developing the evaluation benchmarks for image and video modalities, we adhere to the following core principles:

\begin{itemize}[leftmargin=*, noitemsep]
\renewcommand\labelitemi{$\diamond$}  
    \item \textbf{Targeted capabilities and business relevance:} The internal benchmark is oriented towards measuring the foundational model's ability to handle real user interaction tasks within a \textit{native Chinese context}. We particularly focus on the model's performance in \textit{open-ended question answering} scenarios that require generating complete and valuable responses, in order to assess its generalization capabilities in real business contexts more accurately. In detail, the benchmark not only covers general-purpose capabilities (such as perception and reasoning) but also evaluates potential performance in business scenarios rarely addressed by current open-source benchmarks, such as \textit{multimodal understanding in short video communities}. This design aims to comprehensively assess the model's applied value and ensure it can effectively support actual business requirements.

    \item \textbf{Comprehensive and fine-grained capability taxonomy:} We design a \textit{multi-level capability taxonomy} to facilitate the construction of diverse and fine-grained data tasks and to support the evaluation of the model's capabilities at various levels. At the \textit{macro-capability level} (Layer 2), the evaluation framework covers a range of fundamental abilities, including basic visual skills, multimodal understanding tasks, and complex reasoning, while also incorporating safety and robustness. At the \textit{specific capability level} (Layer 3/4), we further refine these dimensions into a comprehensive set of tasks. 

    \item \textbf{Authentic and diverse data coverage:} To ensure the generalizability of our evaluation and mitigate the risk of overfitting, our benchmark emphasizes diversity in both videos and questions. Our data sources are authentic and reliable; the general benchmark utilizes a large volume of diverse and timely real-world image and video data. The scope of our sampled data is extensive, including various image categories (natural, text-based, artificial) and short videos with diverse frame rates, resolutions, and subjects (people, landscapes, objects, IPs), as well as complex cinematographic and motion elements.

    \item \textbf{Mitigation of benchmark contamination:} To build an evaluation benchmark that genuinely reflects model capabilities and is free from contamination, we have adopted several maintenance strategies. We employ methods such as timestamp verification to prevent data leakage and conduct manual, item-by-item quality checks to ensure data accuracy. Furthermore, we establish a dynamic maintenance mechanism featuring regular updates and on-demand supplementation. This involves continuously refreshing redundant data while adding new tasks that target model weaknesses, ensuring the benchmark remains current and capable of capturing the latest model advancements.

    \item \textbf{Hierarchical evaluation methodology:} We employ a multi-level evaluation system to comprehensively assess the model's open-ended question answering capabilities. This system first involves a \textit{Five-Dimensional Scoring} of model responses based on metrics of \textit{correctness, relevance, comprehensiveness, fluency, and creativity}. To facilitate a holistic comparison, we then calculate a \textbf{Composite Score} by applying task-specific weights to these dimensions.
    
\end{itemize}

\subsubsection{Evaluation Metrics and Baselines}

\paragraph{Input and parameters:}  
For all experiments, input videos are uniformly sampled to select 64 frames. We benchmark Keye-VL against several state-of-the-art open-source models with comparable parameter scales: Qwen2.5-VL-7B, InternVL3-8B, and MiMo-VL-7B-RL. Each model is obtained from its official repository and deployed within our internal computing clusters. Notably, Keye and MiMo utilize a \emph{thinking} mode, enabling internal reasoning steps during inference, while Qwen and Intern generate answers directly without such intermediate processes.

\paragraph{Evaluation data format:}  
To conduct a rigorous evaluation, we extract 150 question-answer pairs from the video subset and 150 from the image subset of our internal benchmark. All questions are open-ended to fully exercise the models’ comprehension and generative abilities, avoiding the limitations inherent in multiple-choice or fill-in-the-blank formats.

\paragraph{Evaluation protocol:}  
To ensure an accurate and authentic reflection of model capabilities, we rely exclusively on human-generated evaluation rather than automated metrics. For both video and image question-answering tasks, three independent annotators score each model response based on defined criteria. The individual scores are then aggregated to produce a final score. In cases where annotator disagreement arises, a professional third-party panel is consulted to arbitrate and ensure consistency.

\paragraph{Five-dimensional metrics:}  We adopt a five-dimensional evaluation framework to assess model outputs:

\begin{itemize}[leftmargin=*, noitemsep]
\renewcommand\labelitemi{$\diamond$}  

\item \emph{Correctness} measures the factual accuracy of the model’s response, including whether it correctly interprets visual content and whether any referenced knowledge is factually sound.

\item \emph{Relevance} assesses the degree to which the response directly addresses the user’s query and remains contextually tied to the visual input.

\item \emph{Comprehensiveness} captures two related aspects: (1) For descriptive tasks, it examines the model’s ability to identify and describe key visual elements, including main subjects and finer details; (2) For question-answering tasks, it evaluates whether the answer includes sufficient explanation or reasoning, rather than merely stating a result.

\item \emph{Fluency} evaluates the linguistic quality of the response, focusing on grammatical correctness, coherence, logical flow, and readability, ensuring the output is easy to understand and free of errors.

\item \emph{Creativity} applies primarily to generative or creative tasks, measuring the originality, imagination, and diversity expressed in the response.
\end{itemize}
\paragraph{Overall evaluation metric:}  
To obtain a unified score that reflects task-specific priorities, we combine the five dimensions with weights dependent on the type of task as follows:
\begin{itemize}[leftmargin=*, noitemsep]
\renewcommand\labelitemi{$\diamond$}  

\item For \emph{question-answering} tasks, the importance hierarchy is: Correctness, Relevance, Comprehensiveness, then Fluency; Creativity is excluded given its limited relevance.

\item For \emph{descriptive} tasks, Correctness remains paramount, followed by Comprehensiveness, then Relevance, and finally Fluency; Creativity is again excluded.

\item For \emph{creative generation} tasks, the order shifts to Relevance and Creativity as primary factors, then Correctness and Comprehensiveness, with Fluency last.
\end{itemize}
To maintain a rigorous standard, if any single evaluation dimension receives a score below 3, the overall score is capped at 3 or below. This rule prevents high aggregate scores from masking critical deficiencies in any one dimension, thereby ensuring a balanced and trustworthy assessment of model performance.

\subsubsection{Evaluation Results}
\textbf{Keye-VL-8B achieves top composite scores demonstrating robust video capabilities:} As shown in Table~\ref{tab:multimodal_eval}, {Keye-VL-8B} achieves the highest overall composite score of \textit{3.33} on the video subset evaluation, demonstrating its {robust and comprehensive capabilities}. The model’s outstanding performance is particularly reflected in the dimensions of \textit{Comprehensiveness (4.36)} and \textit{Creativity (3.75)}, where it significantly outperforms all competing models. This highlights our model’s powerful abilities in video understanding, description, question answering, and creative narration. We attribute this advantage primarily to the carefully designed data and training strategies tailored for video tasks. 

\textbf{The model maintains strong competitiveness across core performance dimensions:} Although the model’s \textit{Correctness} score (\textit{3.34}) is slightly lower than that of Qwen2.5-VL-7B, the margin is minimal and it remains within the top tier. Overall, the evaluation results confirm that Keye-VL-8B maintains competitive performance on fundamental dimensions such as fluency and relevance at a level comparable to industry-leading models, while establishing a distinct competitive edge in complex video tasks requiring deep reasoning and divergent thinking. Importantly, the successful application of this multidimensional scoring framework not only validates the \textit{model’s leading position} but also underscores the \textit{unique value of the evaluation methodology} in penetrating surface-level performance metrics to directly assess core capabilities and user experience.

\textbf{Keye-VL-8B effectively transfers and excels in image domain tasks:} {Keye-VL-8B} extends its capabilities from video to the static image domain, achieving the top ranking on the image subset with a composite score of \textit{3.81}. Its core metrics, including \textit{Correctness (4.05)} and \textit{Comprehensiveness (4.49)}, also lead the industry. This indicates the model’s VQA abilities are both reliable and comprehensive. Notably, we observe an intriguing performance profile: unlike other models whose \textit{Creativity scores} generally increase on image tasks, our model exhibits a \textit{inferior performance} in this dimension (\textit{3.69}). We interpret this not as a deficiency but as a reflection of a deliberate design focus. It further emphasizes Keye-VL-8B’s is weighted toward complex dynamic relational reasoning rather than creative generation on static images.

\textbf{The model shows strong foundational strengths in key video capabilities:} To further dissect the model’s specific competencies on video tasks, we perform a fine-grained breakdown of capability dimensions. As detailed in Table \ref{tab:detailed_capability_eval}, Keye-VL-8B attains the highest scores among all evaluated models in three fundamental areas: \textit{Visual Element Recognition (3.89)}, \textit{(Prior Knowledge-Based) Question Answering (2.78)}, and \textit{Robustness (3.50)}. Additionally, it demonstrates competitive strength in \textit{Temporal Information Understanding (2.92)}, closely matching the leading contenders. These results strongly evidence that our model has established a \textit{solid foundation in core perceptual and cognitive pathways for video understanding}, enabling more accurate and reliable parsing of objective video content. In contrast, some competitors show superior performance in reasoning, creative generation, and domain-specific knowledge dimensions that rely more heavily on large language model generation capabilities. We believe these differences reflect divergent development paths: Keye-VL-8B prioritizes ensuring \textit{correct and stable visual world understanding}, thereby laying a firm groundwork for future improvements in trustworthiness and capability ceilings on more complex reasoning and interactive tasks. 

The model also achieves a significant leading edge in \textit{Basic Description (4.0)}, \textit{Visual Recognition (3.97)}, and \textit{Visual Comprehension (3.70)} of the image subtasks.  Particularly remarkable is the model’s decisive lead on the complex task of \textit{Multi-Image Analysis (3.15)}. We attribute this advantage directly to the model’s deep understanding of correlations across multiple frames, developed through extensive training on massive short video data. This ability to handle temporal information has effectively generalized into analyzing spatial and logical relationships across multiple static images. This finding not only validates our technical approach but also reveals the \textit{unique advantage derived from video-based training}, endowing the model with powerful potential for handling complex visual inputs.

\subsubsection{Analysis of Kwai Keye-VL's Limitations}

Although our model demonstrates competitive performance across multiple benchmarks, we have identified several limitations. These limitations not only reveal areas for improvement in the current model but also guide our directions for the next phase of iterative enhancement.

1. \textbf{Core visual perception capability:} The primary challenge lies in the precise recognition of key visual elements in complex scenes. Specifically, the model still exhibits a non-negligible error rate in OCR, especially for Chinese characters, when processing images containing dense or stylized text. Furthermore, for fine-grained recognition tasks—such as accurately distinguishing specific species of animals or plants, identifying detailed clothing items on persons, or differentiating subtle variations of objects—the model occasionally confuses or misidentifies targets. Regarding information completeness, the model sometimes omits secondary objects or partial textual information within images, indicating that its ability to build a comprehensive understanding of the scene needs strengthening.

2. \textbf{Temporal understanding in video processing:} Our analysis indicates that the model exhibits instability when describing coherent temporal action sequences, particularly in distinguishing between coarse-grained and fine-grained semantic levels of actions. Additionally, the model's perception of cinematic language, such as camera movements and viewpoint shifts, remains relatively weak. It also shows potential for improvement in tasks requiring precise localization of events within video timelines, temporal ordering, tracking object changes, and motion trajectories. These shortcomings constrain the model’s ability to perform deep and coherent analysis of video content.

3. \textbf{Higher-order cognitive and reasoning capability:} The model's reliability declines on problems that demand rigorous logical chains or mathematical calculations. In question involving specialized domain knowledge (e.g., professional disciplines, well-known intellectual properties), the model occasionally produces factual inaccuracies or omissions. Furthermore, although the model performs well in creative generation generally, in scenarios requiring high degrees of originality or deep conceptualization, its outputs may tend toward generic or patterned responses.

\subsection{Quantitative Results}

We provide a selection of qualitative examples illustrating the capabilities of the Kwai Keye-VL model in various aspects. Our presentation is divided into two dimensions: Modality and Thinking Mode. In \Cref{modality_cases}, we demonstrate the Kwai Keye-VL model's capabilities in pure text (\Cref{text_cases}), images (\Cref{image_cases}), and videos (\Cref{video_cases}), including public benchmark cases and business scenario cases. In \Cref{mode_cases}, we showcase the two thinking modes of the Kwai Keye-VL model: Agentic Thinking (\Cref{agentic_cases}) and Auto Thinking (\Cref{auto_cases}).

\section{Discussion}

\subsection{Mutual Enhancement between Reasoning and Non-Reasoning Data}

As discussed in \Cref{cot_mix_cold_start}, following the non-reasoning stage, we introduce the CoT-Mix dataset to cold-start Keye-VL's reasoning capabilities. During this phase, Keye-VL acquires proficiency in both the Non-Thinking and Thinking modes. We evaluate the impact of Step II.1 (cold start with CoT-Mix data) by comparing Keye-VL's performance before and after this training stage. In the Non-Thinking and Thinking modes, Keye-VL's performance improves by 5.67\% and 8.22\% on MMMU, and by 2.95\% and 7.97\% on HallucinationBench, respectively. These results demonstrate that training on CoT-Mix data strengthens both perceptual and reasoning abilities. Furthermore, since similar non-reasoning data is also employed during Steps I.1 and I.2, the observed performance gains in the Non-Thinking mode are primarily attributable to the inclusion of long CoT reasoning data. This suggests that the integration of long CoT data also enhances the model's capability in the Non-Thinking mode, which aligns with recent findings reported in ERNIE 4.5 (~\cite{ernie}).

On MathVista, Keye-VL's accuracy increases by -1.1\% (Non-Thinking) and 7.9\% (Thinking), suggesting that the model spontaneously establishes the connection between the Thinking mode and complex logical reasoning.
Based on this observation, we introduce Auto-Think mode in step II.2 to further enhance Keye-VL's ability to select reasoning modes based on difficulty and improve overall performance.

\subsection{Performance Gain from RL Training}
How to synchronously and stably improve the model's performance across various tasks is one of the core challenges of RL. In Step II.2 (\Cref{mix_rl}), we attempt to achieve this by adopting the Mix-Mode strategy. In our conception, Non-Thinking and Thinking Modes are respectively suitable for simple and complex tasks. Therefore, during the RL process, we apply Keye-VL to generate samples using a mix of both two modes to comprehensively enhance its performance across various tasks, and we incorporate the Auto-Think mode as we discussed before.

The evaluation results show that after RL training, Keye-VL achieves an average improvement of 1.44\%/2.17\% in Non-Thinking/Thinking mode across 10 benchmarks. Its performance increase by 0\%/0.93\% to 4.18\%/5.73\% on 9 benchmarks, with a decrease of only 1.11\%/1.2\% on MMMU/OCRBench. This indicates that mix-mode RL strategy achieves a comprehensive and synchronous improvement.

\subsection{Analysis about Auto-Think Mode}

\begin{table}[h]
    \centering
    \begin{tabular}{ccccc}
        \toprule
        Benchmark & ${\text{MathVista}_\text{MINI}}$ & MMStar  & HallusionBench& OCRBench \\ \midrule
        Thinking Ratio &  0.35 & 0.34& 0.08 & 0.00 \\ \bottomrule
    \end{tabular}
\caption{Proportions of Keye-VL selecting the Thinking mode rather than the Non-Thinking mode across various benchmarks when operating in Auto-Think mode.}\label{tab:auto_think}
\end{table}

To evaluate Keye-VL's ability to spontaneously select the reasoning mode by task type and difficulty, we calculate the proportions of Keye-VL choosing the Thinking mode during Auto-Think inference across four different benchmarks (\cref{tab:auto_think}). 
It turns out that on the more challenging and logic-reasoning-oriented MAthVista and MMStar, Keye more frequently selects the Thinking mode. However, Non-Thinking remains the predominant mode, as the difficulty of most samples is limited. On the HallusionBench, Keye-VL selects the Thinking mode only in few instances. In these cases, Keye-VL engages in reflection to prevent errors, as demonstrated in \cref{image_cases}. On the OCRBench, Keye-VL adopts the Non-Thinking mode for all instances.

\section{Conclusion and Future Work}
In this work, we present Kwai Keye-VL, a model that achieves leading video understanding capabilities through high-quality video data construction during both pre-training and post-training phases. The mix-mode training in the post-training phase further enables Keye-VL to respond more flexibly, resulting in superior user experience. Public benchmarks and internal human evaluations demonstrate Keye-VL's strong performance in general image-text understanding, logical reasoning, and short video applications.

Nevertheless, Keye-VL has room for improvement. First, we have not specifically optimized the video encoder architecture or video encoding strategies, leaving significant potential for future enhancements. Second, Keye-VL's perceptual capabilities show only modest advantages compared to SOTA models, and its ``think with image'' ability remains preliminary, still lagging behind OpenAI's O3 model. Finally, our use of an additional MLLM as a direct reward model introduces limitations in reliability and usability due to the MLLM's inherent instruction-following capabilities and computational costs. Developing more reliable and efficient reward modeling strategies remains an open question for future research.

\bibliography{colm2024_conference}
\bibliographystyle{colm2024_conference}

\newpage
\appendix
\section{Strategies for Data Decontamination} \label{post_decont}

\subsection{Pre-training}
To prevent benchmark leakage (i.e., the model memorizes evaluation examples), we conduct rigorous data decontamination strategies to avoid the data leakage problem.
In pre-training, considering the large-scale data volume, we utilize simple-yet-effective pHash\&minHash techniques to check our data corpus.
We apply a decontamination process based on pHash and minHash. First, we utilize pHash to generate a 64-bit binary string, then convert its ones' ID location list to build minHash index via 128-bit permutation hashing to group them. Next, the same buckets' candidate pairs are verified using Jaccard similarity (>0.95) to identify complex variations like partial crops or local edits.
Considering that some training samples have multiple images, but as long as one image is repeated with the benchmark, we will also filter out the sample from our training data.
In Table~\ref{tab:dedup}, we have listed some open-source datasets that may have data leaks.

\begin{table}[h]
\centering
\footnotesize
\begin{threeparttable}
\setlength{\tabcolsep}{1.5mm}{
\begin{tabular}{lcccccc} \toprule
    Data Source & MMBench & MMMU & AI2D & MMStar & MathVista & OCRbench \\ 
    \midrule
    Infinity Onevision (~\cite{gu2024infinity}) & 3459 & 331 & 902 & 3056 & 2827 & 826 \\
    MMInstruct-QA (~\cite{liu2024mminstruct}) & 1349 & 17 & 0 & 265 & 0 & 1 \\
    MMEvol (~\cite{luo2024mmevol}) & 1842 & 140 & 645 & 1041 & 750 & 380 \\
    \midrule
\end{tabular}}
\end{threeparttable}
\caption{{The number of duplicated pre-training samples with some datasets/benchmarks.}}
\label{tab:dedup}
\end{table}

\subsection{Post Training}
In post-training, we remove any training image-question pair whose visual and textual embeddings are highly similar to instances in our evaluation benchmarks. For each candidate image-question pair in the training corpus, we compute CLIP-based cosine similarities against all evaluation benchmarks using both visual and textual encoders. Pairs exceeding similarity thresholds of 0.98 (image) and 0.50 (text) to any benchmark sample are excluded, preserving semantic diversity while systematically removing evaluation look-alikes.

This decontamination protocol is applied across 29 comprehensive benchmarks spanning five critical evaluation domains: Visual Mathematics \& Reasoning, General Multimodal Understanding, Chart/Diagram Interpretation, OCR-Centric \& Document Tasks, and Robustness \& Hallucination Diagnostics.
The dual-threshold approach effectively safeguards evaluation integrity across all domains while maintaining the original training corpus, ensuring no meaningful reduction in model knowledge coverage.

Meanwhile, Table~\ref{tab:similarity-exceeded-datasets} lists datasets containing a substantial number of samples exceeding the defined thresholds relative to our evaluation benchmarks. Researchers should exercise caution when employing these datasets during model training to avoid potential data leakage.

\begin{table}[htbp]
\centering
\begin{tabular}{lrrrrrr}
\toprule
Dataset     & MMBench & MMMU & AI2D & MMStar & MathVista & OCRBench \\
\midrule
MM-Eureka~(\cite{mm_eureka})   & 2,127   & 0    & 1    & 596    & 616       & 110      \\
MMPR~(\cite{mpo})        & 10,947  & 0    & 1,186 & 2,009  & 3,143     & 1,044    \\
\bottomrule
\end{tabular}
\caption{{The number of duplicated post-training samples between various datasets.}}
\label{tab:similarity-exceeded-datasets}
\end{table}

\section{Construction of KC-MMbench}\label{app_kcmmbench}

The benchmark comprises the following tasks, designed to evaluate short-form video understanding from various perspectives:

\begin{itemize}
    \item \textbf{Standard Product Unit (SPU):} This task involves comparative analysis. We randomly sample information from different products within the same commercial category. The model is then tasked with determining whether two given pieces of product information refer to the same underlying product.

    \item \textbf{Category Property Value (CPV):} We collect data across various product dimensions (e.g., color, style, material). For each dimension, we construct multiple-choice questions using attributes that are semantically close, testing the model's ability to make fine-grained distinctions.

    \item \textbf{Hot Videos Aggregation:} Short videos are collected based on real-world events. A set of videos is constructed through random sampling. One video is designated as an anchor, and the task is to identify which of the other videos in the set belong to the same event as the anchor.

    \item \textbf{Collection Order:} This task uses a collection of topically related short videos published from the same user as input. The objective is to determine the correct logical or chronological sequence of the video content.

    \item \textbf{High-Like Video Classification:} We collect data on the number of "likes" a video receives within a specific timeframe after being uploaded. Using a predetermined threshold, we classify videos into two categories: "high-like" and "low-like".

    \item \textbf{Pornographic Comment:} Leveraging historical data from both automated online moderation models and human reviewers, we randomly sample video information along with associated user comments. This data is then used to formulate a binary classification task to determine whether a given comment contains pornographic content.
\end{itemize}

All datasets sampled for the tasks described above have undergone a thorough manual review process. During this stage, we filter out data with low video quality and verify the correctness of the corresponding ground-truth answers. To protect user privacy, we have meticulously anonymized the data. All personally identifiable information, such as usernames, locations, and timestamps, has been removed from the textual data. For visual data, any sensitive information appearing in images has been blurred to prevent potential information leakage. The final evaluation benchmark is based exclusively on this manually verified and fully anonymized dataset.

\newpage
\begin{longtable}{
    >{\RaggedRight\arraybackslash}p{2.2cm} 
    >{\RaggedRight\arraybackslash}p{3cm} 
    >{\RaggedRight\arraybackslash}p{4.5cm} 
    >{\RaggedRight\arraybackslash}p{3.5cm}
}
\caption{Capability Taxonomy for the Internal General Benchmark.} \label{tab:general_benchmark} \\
\toprule
\textbf{Level 1} & \textbf{Level 2} & \textbf{Level 3} & \textbf{Level 4} \\
\midrule
\endfirsthead
\multicolumn{4}{l}{{\bfseries Table \thetable\ continued from previous page}} \\
\toprule
\textbf{Level 1} & \textbf{Level 2} & \textbf{Level 3} & \textbf{Level 4} \\
\midrule
\endhead
\bottomrule
\endfoot
\multirow{57}{2.2cm}{General Benchmark}
  & \multirow{9}{3cm}{Visual Element Recognition} & Object Recognition & \\
  & & Action Recognition & \\
  & & Attribute Recognition & \\
  & & Spatial Recognition & \\
  & & Temporal Recognition & \\
  & & Location Recognition & \\
  & & Scene/Plot Recognition & \\
  & & OCR Problems & \\
  & & Counting Problems & \\
\cmidrule{2-4}
  & \multirow{9}{3cm}{Temporal Information Understanding} & \multirow{2}{4.5cm}{Camera Motion Changes} & Camera Movement \\
  & & & Shooting Angle Changes \\
\cmidrule{3-4}
  & & \multirow{3}{4.5cm}{Subject Dynamic Perception} & Motion Direction Change \\
  & & & Motion Speed Change \\
  & & & Motion Process Perception \\
\cmidrule{3-4}
  & & \multirow{2}{4.5cm}{State Transition Perception} & Size/Shape/Color/Light Changes \\
  & & & Comprehensive Changes (Duration) \\
\cmidrule{3-4}
  & & \multirow{2}{4.5cm}{Event Temporal Perception} & Event Time Localization \\
  & & & Event Sequence Rearrangement \\
\cmidrule{2-4}
  & \multirow{6}{3cm}{Descriptive Capability} & Video Specific Moment Description & \\
  & & Video Specific Object Description & \\
  & & Video Specific Sub-scene Description & \\
  & & Video Dense Captioning & (Generate description for each event) \\
  & & Summarization of Video & (Major scenes, events, themes) \\
  & & Abstractive Summarization & (For the entire video) \\
\cmidrule{2-4}
  & \multirow{8}{3cm}{Creative Capability} & Basic Writing & (Email, Essay, Story, Novel) \\
  & & Copywriting / Slogan & \\
  & & Poetry & \\
  & & Brainstorming & \\
  & & \multirow{4}{4.5cm}{Multi-turn Conversation} & Conversational Ability \\
  & & & Interactional Intent Understanding \\
  & & & Contextual Understanding \\
  & & & Answer Suggestion and Correction \\
\cmidrule{2-4}
  & \multirow{7}{3cm}{Knowledge-based Q\&A (Prior Knowledge)} & \multirow{5}{4.5cm}{Common Sense Q\&A} & Object Grounding and Naming \\
  & & & Flora/Fauna, Celebrity, Food, Landmark, IP Recognition \\
  & & & General Life Knowledge \\
\cmidrule{3-4}
  & & Music/Video Appreciation & \\
  & & Sports/Gaming Information & \\
\cmidrule{2-4}
  & \multirow{7}{3cm}{Reasoning Capability} & Descriptive Reasoning & \\
  & & Explanatory Reasoning & \\
  & & Comparative Reasoning & \\
  & & Predictive Reasoning & \\
  & & Counterfactual Reasoning & \\
  & & Spatial Relationship Judgment & \\
  & & Causal Relationship Judgment & \\
\cmidrule{2-4}
  & \multirow{5}{3cm}{Domain-Specific Capability} & Chart and Graph Analysis & \\
  & & Academic Problem Solving & \\
  & & Scientific Experiment Analysis & \\
  & & Remote Sensing Video Understanding & \\
  & & Medical Video Understanding & \\
\cmidrule{2-4}
  & \multirow{2}{3cm}{Safety} & Traditional Safety & \\
  & & Instruction Attacks & \\
\cmidrule{2-4}
  & \multirow{4}{3cm}{Robustness} & Video Noise Tests (e.g., detail blur) & \\
  & & Prompt Robustness (typos, unclear intent) & \\
  & & Robustness to varied bitrate, resolution, clarity & \\
  & & Video-Text Irrelevance & \\
\end{longtable}

\newpage

\section{Case Study}

\subsection{Modality}
\label{modality_cases}
\subsubsection{Pure Text Case}
\label{text_cases}
\renewcommand{\arraystretch}{1.5}
\begin{figure}[!h]
  \centering
  \begin{tabular}{m{16cm}}
  \toprule
  {\footnotesize \textbf{Prompts:} "What movie does this emoji describe?
  \includegraphics[height=1em]{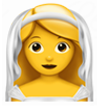} 
  \includegraphics[height=1em]{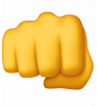} 
  \includegraphics[height=1em]{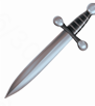} 
  "\newline\newline
  \textbf{Model Responses:} 
  \texttt{<think>}  
  The user now needs to determine which movie the emoji 
  \includegraphics[height=1em]{arXiv/cases/figs/emoji1.png} 
  \includegraphics[height=1em]{arXiv/cases/figs/emoji2.png} 
  \includegraphics[height=1em]{arXiv/cases/figs/emoji3.png} 
  describes. First, analyze the meaning of each emoji: 
  \includegraphics[height=1em]{arXiv/cases/figs/emoji1.png} 
  is a bride,  
  \includegraphics[height=1em]{arXiv/cases/figs/emoji3.png} 
  is a dagger, and  
  \includegraphics[height=1em]{arXiv/cases/figs/emoji2.png} 
  is a fist. 
  
  Then, recall movies that involve a female protagonist in a wedding scene, armed with weapons, and fighting. For example, "Kill Bill: Volume 1" is a Quentin Tarantino film where the main character, Beatrix Kiddo (the Bride), is a bride at her wedding, then seeks revenge, using weapons like a dagger for combat. The elements of the bride, dagger, and fist (representing fighting) all match. Therefore, it should be this movie.  
  \texttt{</think>}  

\texttt{<answer>}  
To determine the movie corresponding to this emoji, it is necessary to analyze the meaning of the emojis and match the plot elements:
  \includegraphics[height=1em]{arXiv/cases/figs/emoji1.png} 
(bride): Indicates a wedding scene;
  \includegraphics[height=1em]{arXiv/cases/figs/emoji3.png} 
(dagger): Represents a weapon;
  \includegraphics[height=1em]{arXiv/cases/figs/emoji2.png} 
(fist): Symbolizes fighting or combat.

Combining these elements, the movie "Kill Bill: Volume 1" fits the plot: the protagonist Beatrix Kiddo (the Bride) is a bride at her wedding, then embarks on a revenge journey, using weapons (such as a dagger) for combat, with the core elements of "bride + weapon + fighting" perfectly matching.

\boxed{\text{Kill Bill: Volume 1}}
\texttt{</answer>}  
}
  \\
  \bottomrule
  {\footnotesize \textbf{Prompts:} "What movie does this emoji describe?
  \includegraphics[height=1em]{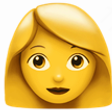} 
  \includegraphics[height=1em]{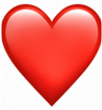} 
  \includegraphics[height=1em]{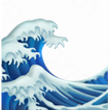} 
  \includegraphics[height=1em]{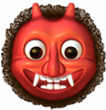} 
  "\newline\newline
  \textbf{Model Responses:} 
  \texttt{<think>}  The user now needs to determine which movie the emoji "
  \includegraphics[height=1em]{arXiv/cases/figs/emj4.png} 
  \includegraphics[height=1em]{arXiv/cases/figs/emj5.png} 
  \includegraphics[height=1em]{arXiv/cases/figs/emj6.png} 
  \includegraphics[height=1em]{arXiv/cases/figs/emj7.png}" describes. First, analyze each element: a woman (  \includegraphics[height=1em]{arXiv/cases/figs/emj4.png} ), a heart ( \includegraphics[height=1em]{arXiv/cases/figs/emj5.png} ), waves (\includegraphics[height=1em]{arXiv/cases/figs/emj6.png} ), and a demon (\includegraphics[height=1em]{arXiv/cases/figs/emj7.png} ). Recall movies with similar elements, such as "The Shape of Water" (The Shape of Water). In the movie, a mute female cleaner falls in love with a mysterious aquatic creature (demon-like?), and the scenes include water and emotional elements. Therefore, it is necessary to confirm the elements corresponding to each part: the female protagonist, the heart symbolizing love, the waves representing the aquatic creature's environment, and the demon-like creature. Thus, the movie is "The Shape of Water."\texttt{</think>}
  
\texttt{<answer>}
To determine the movie corresponding to the emoji "
  \includegraphics[height=1em]{arXiv/cases/figs/emj4.png} 
  \includegraphics[height=1em]{arXiv/cases/figs/emj5.png} 
  \includegraphics[height=1em]{arXiv/cases/figs/emj6.png} 
  \includegraphics[height=1em]{arXiv/cases/figs/emj7.png}", it is necessary to analyze the symbolic meaning of each element:
  \includegraphics[height=1em]{arXiv/cases/figs/emj4.png} : Represents a female character;
  \includegraphics[height=1em]{arXiv/cases/figs/emj5.png} : Symbolizes love or emotional connection;
  \includegraphics[height=1em]{arXiv/cases/figs/emj6.png} : Represents water or an aquatic environment;
  \includegraphics[height=1em]{arXiv/cases/figs/emj7.png} : Can be associated with a mysterious, otherworldly creature (such as an aquatic monster).
Combining these elements, the movie "The Shape of Water" (The Shape of Water) fits well: the story revolves around a mute female cleaner who develops a profound emotional bond with a captured aquatic creature, set against a backdrop filled with water and mysterious creatures.

\boxed{\text{The Shape of Water}}
\texttt{</answer>}  
}
  \\
  \bottomrule
  \end{tabular}
\caption{An example of Kwai Keye-VL successfully deducing the corresponding movie name based on the input emojis. We directly input the emojis in text form and Kwai Keye-VL can accurately understand the meaning of emojis, which is crucial for understanding content on media platforms.}
\label{fig:kill_bill}
\end{figure}
\renewcommand{\arraystretch}{1}

\clearpage

\clearpage
\subsubsection{Image Cases}
\subsubsubsection{Image Case I: Subway Map Puzzle}
\label{image_cases}
\renewcommand{\arraystretch}{1.5}
\begin{figure}[!h]
  \centering
  \begin{tabular}{m{16cm}}
  \toprule
  \begin{center}
  Input Image
  \end{center}
  \begin{center}
  \includegraphics[width=0.5\linewidth]{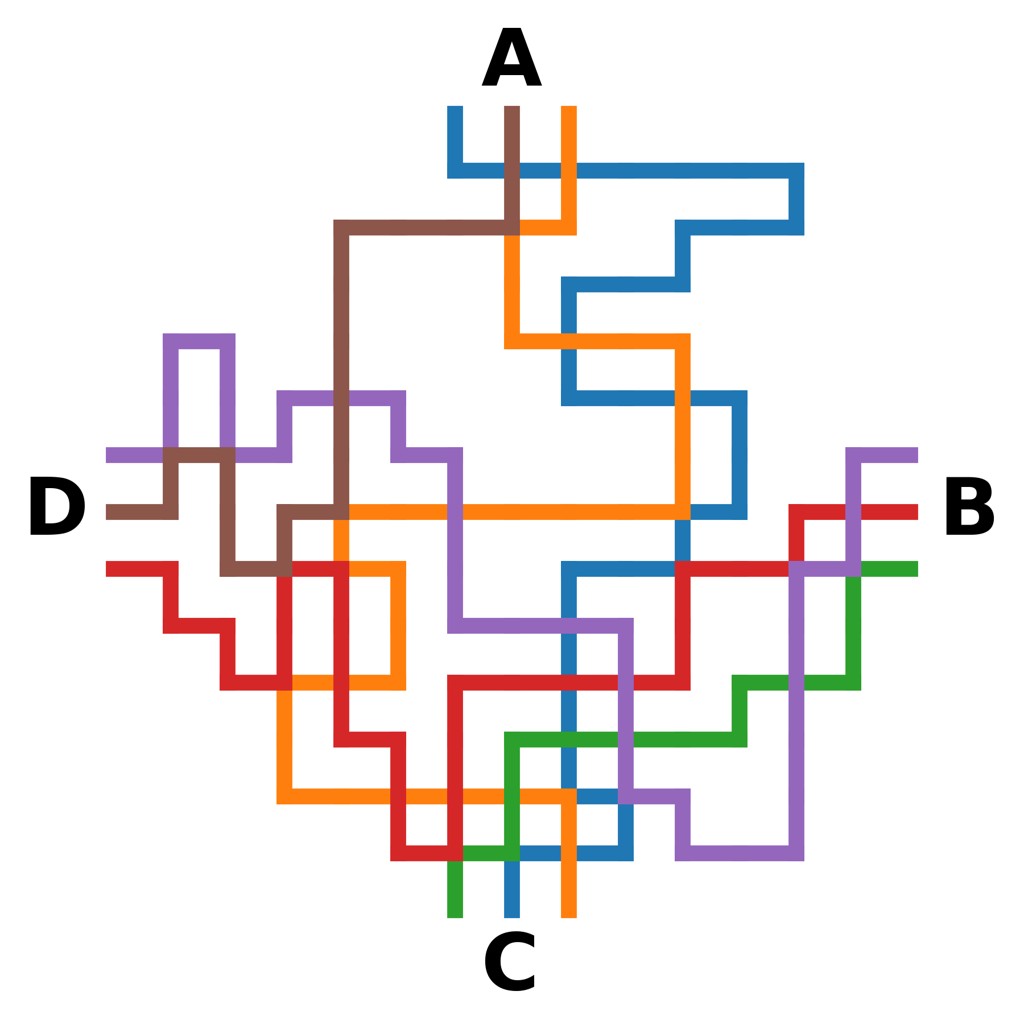}
  \end{center}
  \\
  {\footnotesize \textbf{Prompts:} Count the one-colored routes that go from D to A. Answer with a number in curly brackets e.g. {3}\newline\newline
  \textbf{Model Responses:} \texttt{<think>}The current task for the user is to count the number of single-colored paths from D to A. First, it is necessary to identify the color of each path and then analyze the paths for each color separately to see if they connect D to A.

First, observe the colors of the paths in the image: purple, brown, blue, orange, red, green. Then, check if the paths for each color connect D to A.

Purple path: Check if the purple path connects D to A. Trace the purple path to see if it reaches A. It appears that the purple path does not connect D to A, so it is excluded.

Brown path: Trace the brown path to see if it connects D to A. The brown path seems to connect D to A, so this is one.
Blue path: ... \newline
Orange path: ... \newline
Red path: ... \newline
Green path: ... \newline
Now, check carefully again. Could there be other colors? For example, are there any other colors? Check the colors in the image again: purple, brown, blue, orange, red, green, and possibly others? No, the colors in the image should be these. Then, confirm again if the brown path is the only one? Or are there any others?

Look carefully again, could there be other paths? For example, are there any other colors? Oh, maybe I missed something. Check again:
\newline
  ... \newline
Oh, maybe there are no other colors that connect D to A, so the count is 1?\texttt{</think>}

\texttt{<answer>}To determine the number of single-colored paths from D to A, it is necessary to analyze the connectivity of each colored path one by one:

Purple path: Does not connect D to A, excluded.
Brown path: Continuously connects D to A, meeting the condition.
Blue path: Does not connect D to A, excluded.
Orange path: Does not connect D to A, excluded.
Red path: Does not connect D to A, excluded.
Green path: Does not connect D to A, excluded.
After analyzing each color, only the brown path meets the condition of "single-colored path from D to A."

\boxed{1}

\texttt{</answer>}
}
  \\
  \bottomrule
  \end{tabular}
\caption{An example demonstrating Kwai Keye-VL's reflective ability learned from the RL process. Kwai Keye-VL checks the correctness of the results multiple times before providing the final answer, avoiding misinterpretation of complex images.}
\label{fig:vlm_are_blind_case}
\end{figure}
\renewcommand{\arraystretch}{1}

\clearpage

\clearpage
\subsubsubsection{Image Case II: Comparative Product Verification}
\renewcommand{\arraystretch}{1.5}
\begin{figure}[!h]
  \centering
  \begin{tabular}{m{16cm}}
  \toprule
  \begin{center}
  Input Image
  \end{center}
  \begin{center}
  \includegraphics[width=0.5\linewidth]{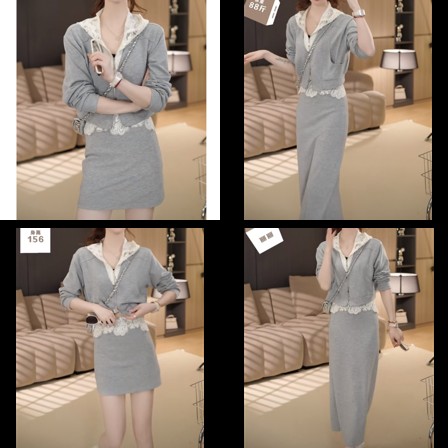}
  \end{center}
  \\
  {\footnotesize \textbf{Prompts:} You are an e-commerce AI assistant with multiple capabilities related to e-commerce operations.

There is a current attribute prediction task where you need to summarize the attributes of a product from specified dimensions based on the main product image and information provided (product title and product description).

Below is the relevant information for the product whose attributes need to be predicted:  
Product Title: New Season Fashionable Lace Hooded Jacket + High Waist Skirt Two-Piece Set LK2501761  
Product Description: New Season Fashionable Lace Hooded Jacket + High Waist Skirt Two-Piece Set LK2501761

Please summarize the product's attribute information from the dimensions of {'Style': ['Skirt Set', 'Strap', 'Fake Two-Piece', 'Pants Set'], 'Clothing Fit': ['Straight', 'Slim', 'Loose', 'Cape Style', 'Batwing', 'Poncho'], 'Collar Type': ['Square Collar', 'Diagonal Collar', 'Scarf Collar', 'Detachable Hood', 'Peter Pan Collar', 'Half-Open Collar', 'Double-Layer Collar', 'Hooded', 'Pile Collar', 'Stand Collar', 'Off-Shoulder', 'Round Collar', 'Half-High Collar', 'High Collar', 'Ruffle Collar', 'V-Neck', 'Polo Collar', 'Sailor Collar', 'Suit Collar', 'None', 'Turn-Down Collar'], 'Sleeve Length': ['Sleeveless', 'Short Sleeve', 'Five-Quarter Sleeve', 'Seven-Quarter Sleeve', 'Nine-Quarter Sleeve', 'Long Sleeve'], 'Applicable Season': ['Spring', 'Summer', 'Autumn', 'Winter', 'All Seasons']} in Chinese. The final attribute summary should be output in JSON string format. Note: 1. If multiple attribute values can be extracted under one attribute dimension, organize them in a list; if only one, do not use a list. 2. If no attributes can be extracted under a specific attribute dimension, do not output that dimension's result. 3. Directly output the final summarized JSON string without additional analysis.\newline\newline
  \textbf{Model Responses:} \texttt{\{"Style": "Skirt Set", "Clothing Fit": "Slim", "Collar Type": "Hooded", "Sleeve Length": "Long Sleeve", "Applicable Season": "Spring"\}
}
}
  \\
  \bottomrule
  \end{tabular}
\caption{An example demonstrating Kwai Keye-VL's application in an e-commerce scenario: Kwai Keye-VL accurately identifies the attributes of a product based on the input product image.}
\label{fig:commercial_case}
\end{figure}
\renewcommand{\arraystretch}{1}

\clearpage

\clearpage
\subsubsection{Video Cases}
\label{video_cases}

\subsubsubsection{Video Case I: Comment Understanding}
\label{case_video_comment}
\renewcommand{\arraystretch}{1.5}
\begin{figure}[!h]
  \centering
  \begin{tabular}{m{16cm}}
  \toprule
  \begin{center}
  Input Video
  \end{center}
  \begin{center}
  \includegraphics[width=0.5\linewidth]{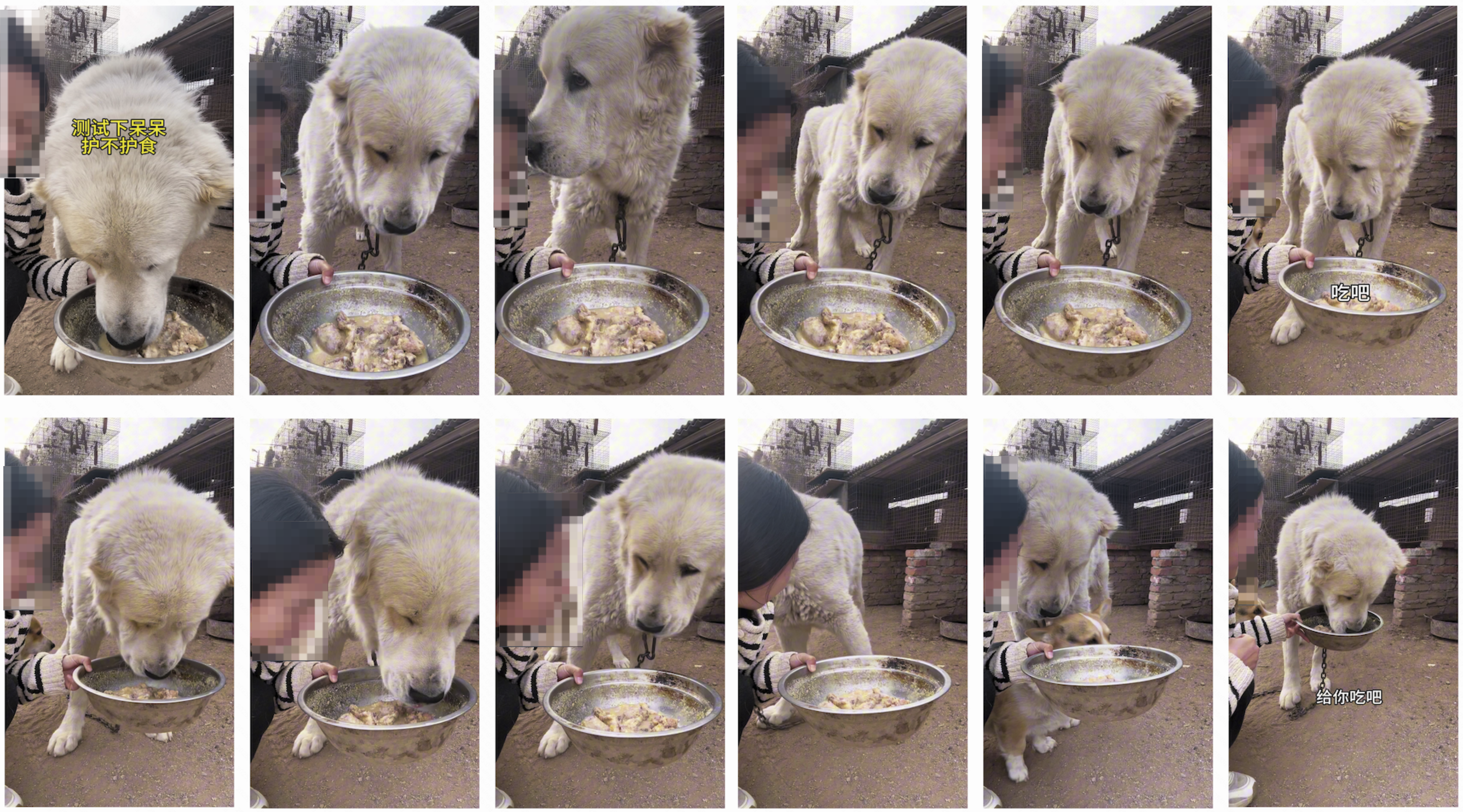}
  \end{center}
  \\
{\footnotesize \textbf{Prompts:} You are an expert in understanding video comments. Please analyze and explain why this comment was rated as a top comment based on the video frames [video] and text.
Video content information: [Video title: [Testing if Dada protects food? \#Dog \#DundunDada]
Video ASR: [Yeah. | Already finished eating. | So cold.]
Video content OCR information: [Testing Dada. Protects food or not | Eat | Give]]
Comment content: [Dada: She can't bite me, right [Oh]];\newline\newline
\textbf{Model Responses:} {Reason Analysis: The reason this comment was rated as a top comment is due to its creativity and humor. The comment uses personification, posing the question "She can't bite me, right" from the perspective of the dog "Dada," creating a clever contrast. In typical food protection tests, people worry about dogs biting to protect their food, but the comment reverses this thinking, making the dog worry about the owner biting it. This reversal breaks conventional thinking and creates a humorous effect. Additionally, the comment vividly showcases the cute interaction between the pet and the owner, aligning with the video's theme of testing food protection. It precisely captures the fun aspect of the video, resonating with the audience and eliciting a smile. The language is concise yet vivid, making it a top comment.}\newline
}
  \\
  \bottomrule
  \end{tabular}
\caption{An example demonstrating Kwai Keye-VL's understanding of short video comments: Kwai Keye-VL accurately identifies the commenter imitating a pet talking to its owner, creating a contrast and humorous effect.}
\label{fig:video_comment}
\end{figure}
\renewcommand{\arraystretch}{1}

\clearpage

\subsubsubsection{Video Case II: Physical Problem Reasoning}
\label{case_physical_video}
\renewcommand{\arraystretch}{1.5}
\begin{figure}[!h]
  \centering
  \begin{tabular}{m{16cm}}
  \toprule
  \begin{center}
  Input Video
  \end{center}
  \begin{center}
  \includegraphics[width=0.9\linewidth]{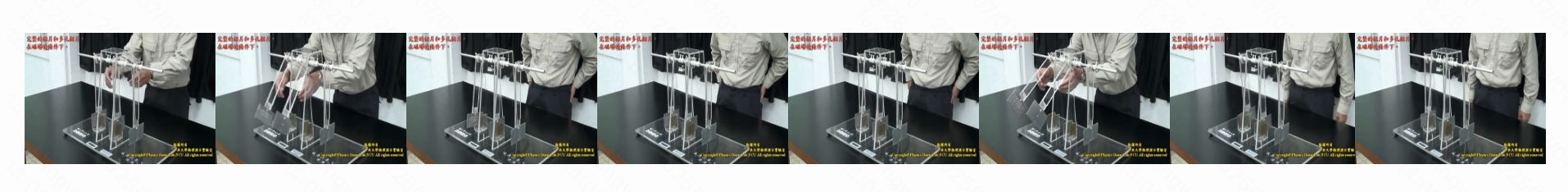}
  \end{center}
  \\
{\footnotesize \textbf{Prompts:} Which of the following statements best explains the observed difference in the durations of swinging, and predicts the expected effect if the middle plate were replaced with a plate made of brass?

A. The middle plate stops sooner because the magnetic field exerts a direct mechanical force opposing its motion; replacing it with brass would cause it to stop even sooner due to increased magnetic interactions.

B. The middle plate stops sooner due to magnetic hysteresis losses as it moves through the magnetic field; replacing it with brass would cause it to stop later because brass is less susceptible to hysteresis losses.

C. The middle plate stops sooner due to eddy current damping opposing its motion; replacing it with brass, which has higher electrical resistivity, would cause it to swing shorter because larger eddy currents would be induced, leading to more damping.

D. The middle plate stops sooner because the magnetic field exerts a direct mechanical force opposing its motion; replacing it with brass would cause it to stop later due to decreased magnetic interactions.

E. The middle plate stops sooner due to eddy current damping opposing its motion; replacing it with brass, which has higher electrical resistivity, would cause it to swing longer because smaller eddy currents would be induced, leading to less damping.\newline\newline
\textbf{Model Responses:} {<think>
Got it, let's try to work through this problem. First, the question is about why the middle plate stops sooner when swinging, and what happens if we replace it with brass. The options mention eddy currents, magnetic hysteresis, direct mechanical force, resistivity, etc. 

First, I need to remember what the setup is. The video shows a pendulum with different metal plates swinging through a magnetic field. The key here is ...

The problem mentions "the middle plate" stops sooner. Let's assume there are ...

Now, the options: Let's go through them. 

Option A: Direct mechanical force. No, magnetic force on the plate isn't direct mechanical; it's due to induced currents. So A is out.

...

Option E: Eddy current damping. Correct mechanism. Middle plate stops sooner due to eddy currents. Then, brass has higher resistivity than, say, copper. Higher resistivity means smaller eddy currents (since eddy current is proportional to 1/resistivity, other factors constant). So smaller eddy currents mean less damping, so the plate swings longer. That matches E.

Let me confirm: The induced current depends on the resistance of the material. For a given induced emf, current I = emf/R. So if R is higher, then I is smaller. ... That's option E.</think>

<answer>To determine the best explanation for the observed difference in swinging durations and the effect of replacing the middle plate with brass, we analyze the phenomenon using principles of electromagnetism, specifically eddy current damping:

\#\#\# **Key Observations and Analysis**  

The setup involves metal plates ...

\#\#\# **Evaluation of Options**  

- **Option A**: Incorrect...

...

- **Option E**: Correct. The middle plate stops sooner due to eddy current damping (larger eddy currents in lower-resistivity material). Brass has higher electrical resistivity than the original material (e.g., copper or aluminum), so smaller eddy currents are induced, reducing damping. This results in the brass plate swinging longer.

**Final Answer** \boxed{\text{E}}</answer>}\newline
}
  \\
  \bottomrule
  \end{tabular}
\caption{An example demonstrating how Kwai Keye-VL solves complex physics problems based on video content. Kwai Keye-VL accurately extracts the objects and setup from the video, analyzes each option using physics knowledge, checks them one by one, and finally confirms the correct answer.}
\label{fig:physical_video}
\end{figure}
\renewcommand{\arraystretch}{1}

\clearpage

\clearpage
\subsection{Thinking Mode}
\label{mode_cases}

\subsubsection{Agentic Thinking Case}
\label{agentic_cases}
\renewcommand{\arraystretch}{1.5}
\begin{figure}[h]
  \centering
  \begin{tabular}{m{6cm}m{10cm}}
  \toprule
  \includegraphics[width=6cm]{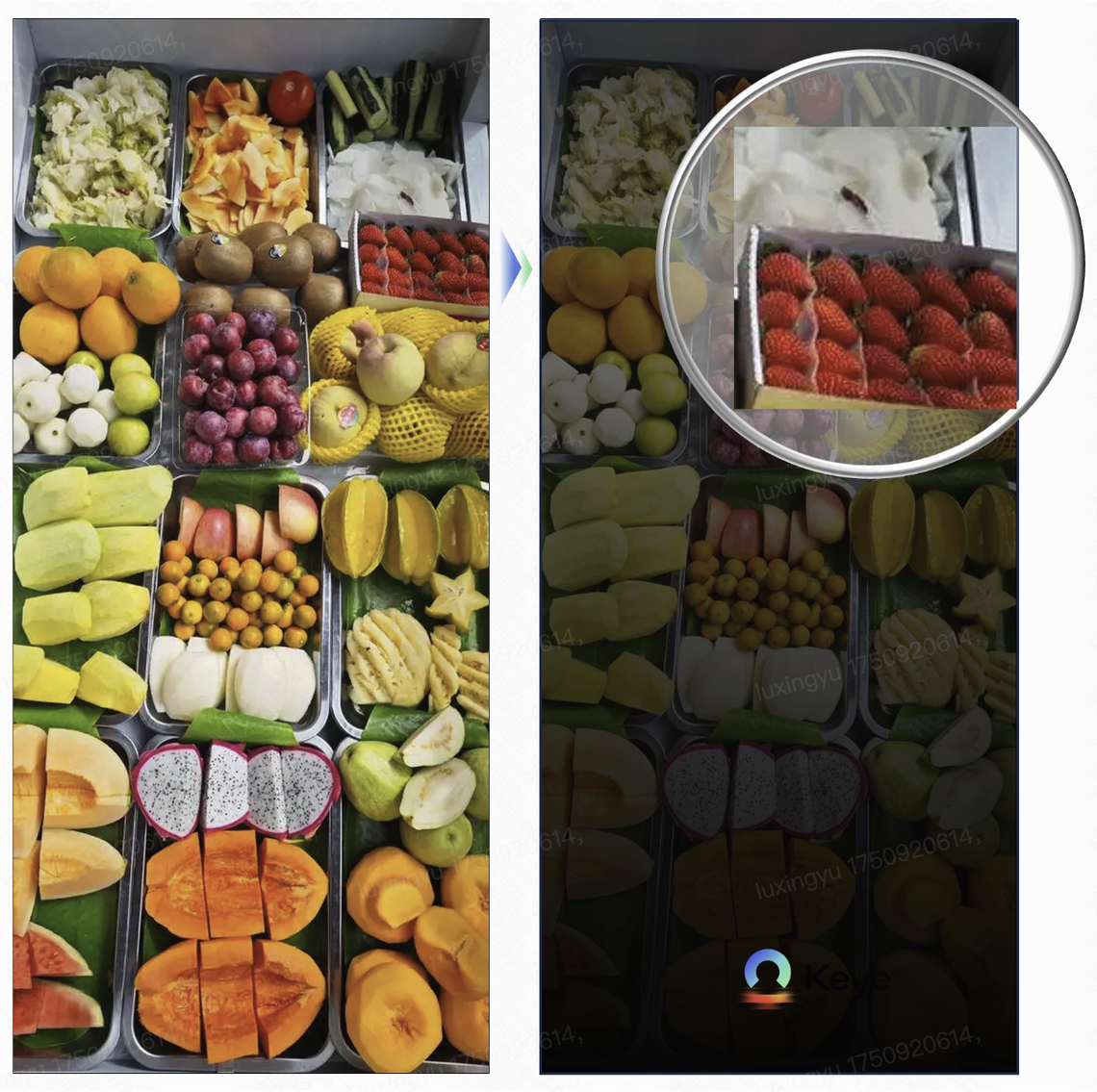} & 
  {\footnotesize \textbf{Prompts:} How many strawberries are there in the picture in total?
\par
  \textbf{Model Responses:} \texttt{<think>}The user is asking about the number of strawberries in the picture. The strawberries are located in the tray at the top right corner of the image. To accurately count the number of strawberries, you can zoom in on this specific area. By cropping and enlarging this area, each strawberry will become clearer, making the count more accurate. The coordinates for this area are approximately (350, 200, 470, 350).
Here is the Python code for cropping and enlarging the strawberry area:
  \par
  \begin{Verbatim}[breaklines=true,breaksymbol=]
    import os
    import random
    from PIL import Image
    
    \# Load the image
    image_path = "test.webp"
    img = Image.open(image_path)
    
    \# Define the coordinates for the strawberry region
    x1, y1, x2, y2 = 350, 150, 500, 300
    
    \# Crop the image
    cropped_img = img.crop((x1, y1, x2, y2))
    
    \# Apply a zoom effect by resizing the cropped image
    zoom_factor = 2
    zoomed_img = cropped_img.resize((cropped_img.width * zoom_factor, cropped_img.height * zoom_factor))
    
    \# Save the processed image with a random suffix
    random_suffix = random.randint(1000, 9999)
    processed_filename = f"test_{random_suffix}.webp"
    processed_path = "/mnt/data/temp_processed_images/" + processed_filename
    zoomed_img.save(processed_path)
    
    \# Print the path of the saved cropped image
    print(processed_path)
  \end{Verbatim}
  \par


<sandbox\_output>
<subimage\_strawberry>
</sandbox\_output>

By carefully examining the cropped and enlarged image, it is clear that the number of strawberries is 20. The enlarged image shows the strawberries more clearly, with no overlapping or blurring, making the count more accurate.
\texttt{</think>}

\texttt{<answer>}
The number of strawberries in the picture is \boxed{20}.
\texttt{</answer>}
} \\

  \bottomrule
  \end{tabular}
\caption{An example of Kwai Keye-VL writing code to extract the relevant parts from high-resolution images and successfully solves the counting problem. In the agentic thinking mode, Kwai Keye-VL can independently write code to edit input images or call tools.
}
\label{fig:strawberry}
\end{figure}
\renewcommand{\arraystretch}{1}

\clearpage

\subsubsection{Auto Thinking Cases}
\label{auto_cases}
\subsubsubsection{No-Thinking Case}
\renewcommand{\arraystretch}{1.5}
\begin{figure}[!h]
  \centering
  \begin{tabular}{m{16cm}}
  \toprule
  \begin{center}
  Input Image
  \end{center}
  \begin{center}
  \includegraphics[width=0.5\linewidth]{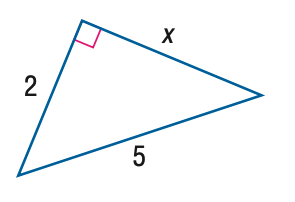}
  \end{center}
  \\
  {\footnotesize \textbf{Prompts:} Find x.\newline\newline
  \textbf{Model Responses:} 
  <analysis>This question requires finding the value of x using the Pythagorean theorem, which is a straightforward calculation given the right triangle. Therefore, /no\_think mode is more appropriate.</analysis>
  
  The given image depicts a right triangle with one leg measuring 2 units, the hypotenuse measuring 5 units, and the other leg labeled as \( x \). To find the value of \( x \), we can use the Pythagorean theorem, which states that in a right triangle, the square of the length of the hypotenuse (\( c \)) is equal to the sum of the squares of the lengths of the other two sides (\( a \) and \( b \)):

\[ c^2 = a^2 + b^2 \]

In this case, the hypotenuse \( c \) is 5, one leg \( a \) is 2, and the other leg \( b \) is \( x \). Plugging these values into the theorem, we get:

\[ 5^2 = 2^2 + x^2 \]

Simplifying the equation:

\[ 25 = 4 + x^2 \]

Subtract 4 from both sides to isolate \( x^2 \):

\[ 21 = x^2 \]

Taking the square root of both sides to solve for \( x \):

\[ x = \sqrt{21} \]

Thus, the value of \( x \) is \( \sqrt{21} \).  
  }
  \\
  \bottomrule
  \end{tabular}
\caption{An example demonstrating Kwai Keye-VL analyzing the difficulty of a problem and choosing the Non-Thinking mode to solve a simple geometry problem.}
\label{fig:no_think}
\end{figure}
\renewcommand{\arraystretch}{1}

\clearpage

\subsubsubsection{Thinking Case}
\renewcommand{\arraystretch}{1.5}
\begin{figure}[!h]
  \centering
  \begin{tabular}{m{16cm}}
  \toprule
  \begin{center}
  Input Image
  \end{center}
  \begin{center}
  \includegraphics[width=0.5\linewidth]{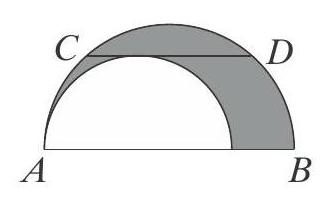}
  \end{center}
  \\
  {\footnotesize \textbf{Prompts:} Find x.\newline\newline
  \textbf{Model Responses:} 
 \texttt{<analysis>}
 This question involves calculating the area of a shaded region formed by two semicircles, where specific geometric properties and relationships between the semicircles and the chord are given. This requires understanding the geometry and performing calculations to determine the area. Therefore, /think mode is more appropriate.\texttt{</analysis>}
 
\texttt{ <think>}The problem that users need to solve now is the area of the shaded region. First, it is necessary to understand the composition of the figure: the shaded region should be the area of the larger semicircle minus the area of the smaller semicircle, or the area of the larger semicircle minus the area of the smaller semicircle plus the area of the smaller semicircle? No, looking at the figure, the shaded region is the part of the larger semicircle excluding the smaller semicircle, so it is the area of the larger semicircle minus the area of the smaller semicircle. Then, it is necessary to find the radii of the larger and smaller semicircles.

First, let the radius of the larger semicircle be R and the radius of the smaller semicircle be r. Since CD is parallel to AB and CD touches the smaller semicircle, the distance from the center O of the larger semicircle to CD is equal to the radius r of the smaller semicircle. The length of CD is 4, so in the right triangle formed by half of CD, the radius R of the larger semicircle, and the distance r, according to the Pythagorean theorem, we have (CD/2)² + r² = R², which is 2² + r² = R², i.e., 4 + r² = R².

Then, the area of the shaded region is the area of the larger semicircle minus the area of the smaller semicircle, which is 
$\frac{1}{2}\pi R^2 - \frac{1}{2}\pi r^2 = \frac{1}{2}\pi (R^2 - r^2)$. From the previous equation,
$R^2 - r^2 = 4$, 
so the area is $\frac{1}{2}\pi \times 4 = 2\pi$. Thus, the area of the shaded region is $2\pi$.\texttt{</think>}

\texttt{<answer>}
To solve this problem, we can analyze the area of the shaded region by combining the Pythagorean theorem and the formula for the area of a circle.

Step 1: Define Variables and Analyze Geometric Relationships
Let the radius of the larger semicircle be ( R ) and the radius of the smaller semicircle be ( r ). Observing the figure, we know that the chord ( CD ) is parallel to the diameter ( AB ) of the larger semicircle and is tangent to the smaller semicircle. Therefore, the distance from the center ( O ) of the larger semicircle to ( CD ) is equal to the radius ( r ) of the smaller semicircle.

The length of ( CD ) is 4, so half of ( CD ) is ( $\frac{CD}{2} = 2$ ). At this point, a right triangle can be formed: one leg is ( $\frac{CD}{2} = 2$ ), the other leg is the distance from the center ( O ) to ( CD ) (i.e., ( r )), and the hypotenuse is the radius ( R ) of the larger semicircle.

Step 2: Use the Pythagorean Theorem to Find ( $R^2 - r^2$ )
According to the Pythagorean theorem, in the right triangle, we have: [ $\left( \frac{CD}{2} \right)^2 + r^2 = R^2$ ] Substituting ( $\frac{CD}{2} = 2$ ), we get: [ $2^2 + r^2 = R^2 \implies 4 + r^2 = R^2 \implies R^2 - r^2 = 4$ ]

Step 3: Calculate the Area of the Shaded Region
The area of the shaded region is equal to the area of the larger semicircle minus the area of the smaller semicircle. According to the formula for the area of a circle ( $S = \pi r^2$ ), the area of the semicircle is ( $\frac{1}{2} \pi r^2$ ). Therefore: [ \text{Area of the shaded region} = $\frac{1}{2} \pi R^2 - \frac{1}{2} \pi r^2 = \frac{1}{2} \pi (R^2 - r^2)$ ] Substituting ( $R^2 - r^2 = 4 $), we get: [ \text{Area of the shaded region} = $\frac{1}{2} \pi \times 4 = 2\pi$ ]

In conclusion, the area of the shaded region is ($\boxed{2\pi}$).
\texttt{</answer>}
  }
  \\
  \bottomrule
  \end{tabular}
\caption{An example demonstrating Kwai Keye-VL choosing the Thinking mode to solve a hard geometry problem.}
\label{fig:think}
\end{figure}
\renewcommand{\arraystretch}{1}

\clearpage

\section{Authors (Alphabetical order)}
\textbf{Core Contributors:} Biao Yang, Bin Wen, Changyi Liu, Chenglong Chu, Chengru Song, Chongling Rao, Chuan Yi, Da Li, Dunju Zang, Fan Yang, Guorui Zhou, Hao Peng, Haojie Ding, Jiaming Huang, Jiangxia Cao, Jiankang Chen, Jingyun Hua, Jin Ouyang, Kaibing Chen, Kaiyu Jiang, Kaiyu Tang, Kun Gai, Shengnan Zhang, Siyang Mao, Sui Huang, Tianke Zhang, Tingting Gao, Wei Chen, Wei Yuan, Xiangyu Wu, Xiao Hu, Xingyu Lu, Yang Zhou, Yi-Fan Zhang, Yiping Yang, Yulong Chen, Zhenhua Wu, Zhenyu Li, Zhixin Ling, Ziming Li.

\textbf{Contributors:} Dehua Ma, Di Xu, Haixuan Gao, Hang Li, Jiawei Guo, Jing Wang, Lejian Ren, Muhao Wei, Qianqian Wang, Qigen Hu, Shiyao Wang, Tao Yu, Xinchen Luo, Yan Li, Yiming Liang, Yuhang Hu, Zeyi Lu, Zhuoran Yang, Zixing Zhang.

\end{document}